\documentclass[letterpaper]{article} 
\usepackage{aaai2026}  
\usepackage{times}  
\usepackage{helvet}  
\usepackage{courier}  
\usepackage[hyphens]{url}  
\usepackage{graphicx} 
\usepackage{natbib}  
\usepackage{caption} 
\usepackage{algorithm}
\usepackage{algorithmic}
\usepackage{amsmath}
\usepackage{amssymb}
\usepackage{siunitx}
\usepackage{booktabs}
\usepackage{comment}
\usepackage{cuted}

\usepackage{newfloat}
\usepackage{listings}
\DeclareCaptionStyle{ruled}{labelfont=normalfont,labelsep=colon,strut=off} 
\floatstyle{ruled}
\newfloat{listing}{tb}{lst}{}
\floatname{listing}{Listing}
\usepackage[colorlinks=true,citecolor=brown,urlcolor=blue]{hyperref}
\usepackage[noabbrev,capitalize]{cleveref}
\crefname{section}{Section}{Sections}
\Crefname{section}{Section}{Sections}
\crefname{subsection}{Section}{Sections}
\Crefname{subsection}{Section}{Sections}
\crefname{subsubsection}{Section}{Sections}
\Crefname{subsubsection}{Section}{Sections}

\usepackage{tcolorbox}
\tcbuselibrary{skins, breakable, theorems}
\usepackage{colortbl}
\usepackage{makecell}
\usepackage{hhline}
\usepackage{array}
\usepackage{subcaption}
\usepackage{multirow}

\usepackage[utf8]{inputenc} 
\usepackage[T1]{fontenc}    
\usepackage{relsize}
\title{DexGraspVLA: A Vision-Language-Action Framework Towards General Dexterous Grasping}
\author {
    \normalsize
    Yifan Zhong\textsuperscript{\rm 1,2}\equalcontrib,
    Xuchuan Huang\textsuperscript{\rm 1,2}\equalcontrib,
    Ruochong Li\textsuperscript{\rm 2,3},
    Ceyao Zhang\textsuperscript{\rm 1,2},
    Zhang Chen\textsuperscript{\rm 1,2},
    Tianrui Guan\textsuperscript{\rm 1,2}\\
    Fanlian Zeng\textsuperscript{\rm 2,4},
    Ka Nam Lui\textsuperscript{\rm 1,2},
    Yuyao Ye\textsuperscript{\rm 1,2},
    Yitao Liang\textsuperscript{\rm 1,2},
    Yaodong Yang\textsuperscript{\rm 1,2$\dag$},
    Yuanpei Chen\textsuperscript{\rm 1,2}\thanks{Corresponding author emails: yuanpei.chen312@gmail.com, yaodong.yang@pku.edu.cn.}
}
\affiliations {
    \normalsize
    \textsuperscript{\rm 1}Institute for Artificial Intelligence, Peking University.
    \textsuperscript{\rm 2}PKU-PsiBot Joint Lab.\\
    \textsuperscript{\rm 3}Hong Kong University of Science and Technology (Guangzhou).
    \textsuperscript{\rm 4}University of Pennsylvania.\\
}

\begin{document}

\maketitle

\begin{abstract}
Dexterous grasping remains a fundamental yet challenging problem in robotics. A general-purpose robot must be capable of grasping diverse objects in arbitrary scenarios. However, existing research typically relies on restrictive assumptions, such as single-object settings or limited environments, showing constrained \emph{generalization}. We present \textbf{DexGraspVLA}, a hierarchical framework for robust generalization in language-guided general \textbf{dex}terous \textbf{grasp}ing and beyond. It utilizes a pre-trained \textbf{V}ision-\textbf{L}anguage model as the high-level planner and learns a diffusion-based low-level \textbf{A}ction controller. The key insight to achieve \emph{generalization} lies in iteratively transforming diverse language and visual inputs into domain-invariant representations via foundation models, where imitation learning can be effectively applied due to the alleviation of domain shift. Notably, our method achieves a \emph{90+\%} dexterous grasping success rate under \emph{thousands of} challenging unseen cluttered scenes. Empirical analysis confirms the \emph{consistency} of internal model behavior across environmental \emph{variations}, validating our design. DexGraspVLA also, for the first time, simultaneously demonstrates free-form long-horizon prompt execution, robustness to adversarial objects and human disturbance, and failure recovery. Extended application to nonprehensile grasping further proves its generality. Project website: \url{https://dexgraspvla.github.io}.
\end{abstract}


\vspace{-12pt}

\section{Introduction}

\begin{figure}[t]
    \centering
    \vspace{-15pt}
    \includegraphics[width=0.95\linewidth]{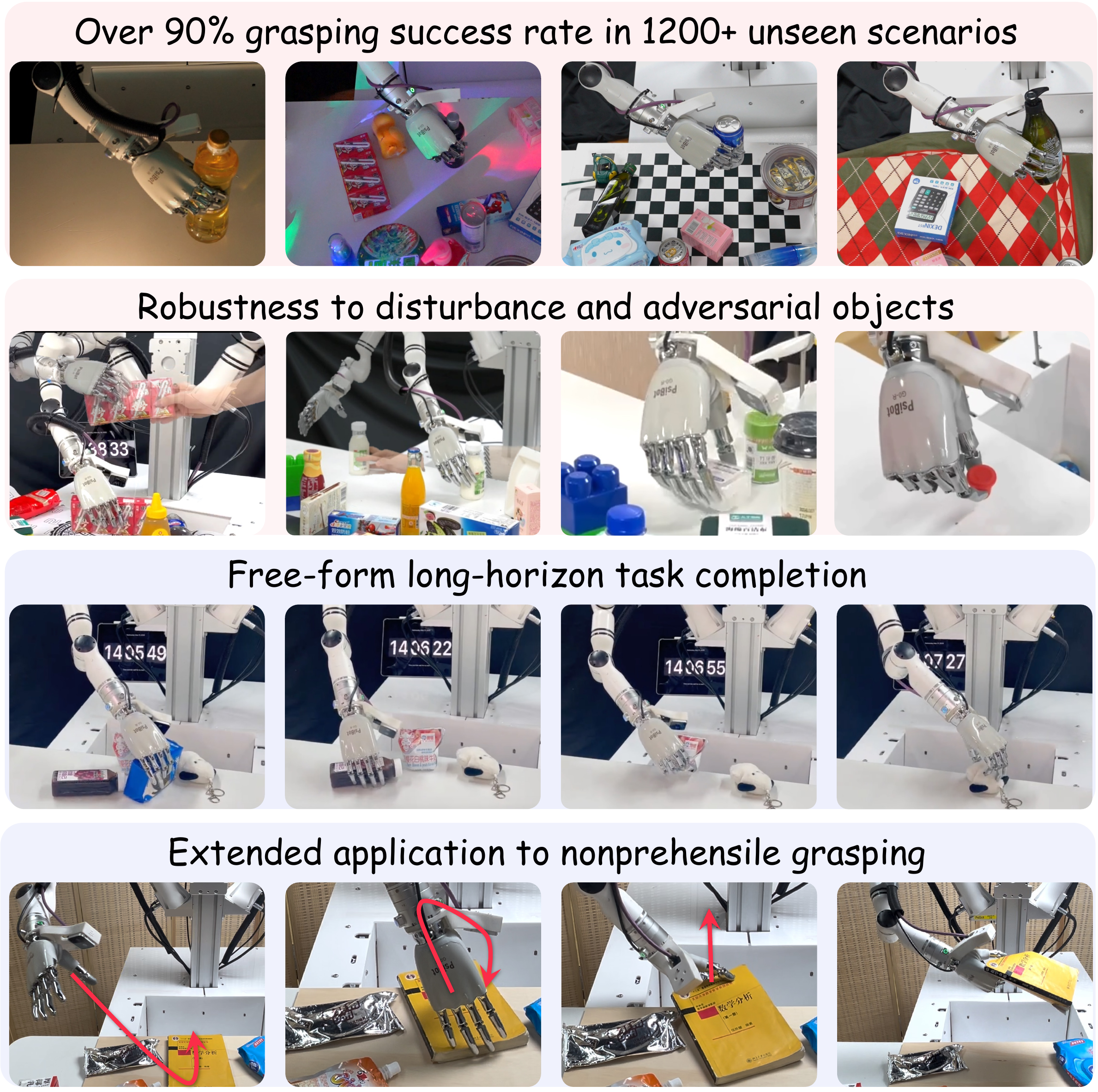} 
    \caption{We propose \textbf{DexGraspVLA}, a hierarchical VLA framework that reaches a 90+\% dexterous grasping success rate under thousands of unseen cluttered scenes in a ``zero-shot'' environment. It robustly handles adversarial objects, human disturbance, failure recovery, and free-form long-horizon prompts. Extended application to nonprehensile grasping further proves its generality.}
    \label{fig:teaser}
    \vspace{-10pt}
\end{figure}

Dexterous multi-fingered hands, as versatile robotic end-effectors, have demonstrated remarkable capabilities across various manipulation tasks~\cite{qi2023general, huang2023dynamic, lin2024twisting, chen2022towards, zakka2023robopianist, chen2023sequential}. Among these, grasping serves as the most fundamental prerequisite, yet remains one of the most challenging problems. Existing dexterous grasping approaches primarily consider isolated objects or simplified settings. Nevertheless, real-world applications demand more general grasping capabilities that can function reliably in diverse unseen scenarios, which presents multifaceted challenges. At the object level, the policy must generalize across diverse physical properties including geometries, masses, textures, and orientations. Beyond object characteristics, the system must also demonstrate robustness to various environmental factors, such as lighting conditions, background complexities, and potential disturbances. Compounding these challenges, cluttered scenarios further demand sophisticated reasoning capabilities, as planning the optimal sequence to grasp all objects becomes a crucial cognitive task that extends beyond simple grasp execution.

One line of research adopts a two-stage pipeline: first predicting a grasp pose from single-frame perception, then executing open-loop motion planning to reach the pose~\cite{chen2024springgrasp, turpin2023fast, turpin2022grasp}. However, these methods rely heavily on precise calibration and mechanical accuracy. By contrast, end-to-end paradigms, such as imitation and reinforcement learning, enable closed-loop control by continuously adjusting actions based on real-time feedback, offering more robust and adaptive solutions. Reinforcement learning has achieved notable successes in simulation~\cite{akkaya2019solving,yang2024anyrotate,pitz2023dextrous,handa2022dextreme}, but simulating real-world physical complexity remains challenging, resulting in an inevitable sim-to-real gap. Imitation learning learns directly from human demonstrations and avoids this gap, but often struggles to generalize beyond the training data. This issue is further compounded by the impracticality of collecting expert trajectories across the full spectrum of objects and environmental variations required for general grasping. As a result, a key challenge is how to effectively leverage limited expert data to achieve broad \emph{generalization}.

The rapid emergence of vision and language foundation models~\cite{oquab2023dinov2,radford2021learning,hurst2024gpt,kirillov2023segment} presents promising opportunities for robotic manipulation. Pretrained on internet-scale data, these models exhibit remarkable world knowledge and generalization over visual and linguistic inputs. To harness these capabilities for decision making, researchers have integrated them into action generation, leading to the development of vision-language-action (VLA) models~\cite{zhong2025survey}. One straightforward approach directly trains vision-language models (VLMs) end-to-end on robot data~\cite{kim2024openvla, black2024pi0}. However, this paradigm demands massive manually collected demonstrations~\cite{o2023open} in an attempt to encompass real-world diversity and complexity. Even so, these models exhibit markedly reduced performance on unseen scenarios and still require fine-tuning to handle new conditions. 
Alternatively, modular frameworks use frozen foundation models to infer task affordances more robustly across environments~\cite{huang2024rekep,huang2023voxposer,stoneopen}, but their low-level policies are typically open-loop or lack generalization. Achieving generalizable closed-loop policies with foundation models remains an open challenge.

In this paper, we present \textbf{DexGraspVLA}, a hierarchical VLA framework for robust generalization in language-guided dexterous grasping and beyond, by integrating the complementary strengths of foundation models and imitation learning. The key idea is to leverage foundation models to iteratively transform \emph{diverse} visual and linguistic inputs into \emph{domain-invariant} representations, upon which imitation learning can be efficiently and effectively applied thanks to the alleviation of \emph{domain shift}. As a result, novel scenarios no longer induce failure, as foundation models translate them into representations resembling those encountered during training—thus remaining within the learned policy’s domain. Following this principle, DexGraspVLA employs a pre-trained VLM as a high-level planner to plan the overall task and generate \emph{domain-invariant} affordance signals. Guided by these signals, a low-level controller further refines multimodal inputs into \emph{domain-invariant} representations using vision foundation models, and generates closed-loop action through a diffusion-based action head learned via imitation. This design combines the extensive world knowledge and generalization ability of foundation models with action modeling capacity of imitation learning, enabling strong performance in real-world scenarios.

Notably, DexGraspVLA achieves an unprecedented 90.8\% success rate for grasping in cluttered scenes spanning 1,287 unseen object, lighting, and background combinations, all tested in a ``zero-shot'' environment. Its generalization performance significantly surpasses that of existing baselines. Moreover, DexGraspVLA robustly handles adversarial objects, human disturbances, and failure recovery. On a single-object benchmark, it achieves 98.6\% success, outperforming ablated variants whose controller learns directly from raw visual inputs by at least 48\%. Further analysis reveals \emph{consistent} internal model behaviours across \emph{varying} environments, validating our design and explaining its robustness. Beyond single-step tasks, DexGraspVLA executes free-form, long-horizon instructions with embodied reasoning, reaching 89.6\% success rate. We further extend DexGraspVLA to nonprehensile object grasping~\cite{zhou2023learning}, a challenging task that often requires dexterous pre-grasp maneuvers difficult for parallel grippers. DexGraspVLA achieves strong performance using only a modest number of demonstrations, further highlighting its generality across diverse manipulation scenarios. These results establish DexGraspVLA as a general, instruction-driven framework that learns from limited demonstrations and generalizes reliably to real-world settings, marking a promising step toward general dexterous grasping and beyond.

\section{Related Work}

\begin{figure*}[t]
    \centering
    \vspace{-5pt}
    \includegraphics[width=0.8\linewidth]{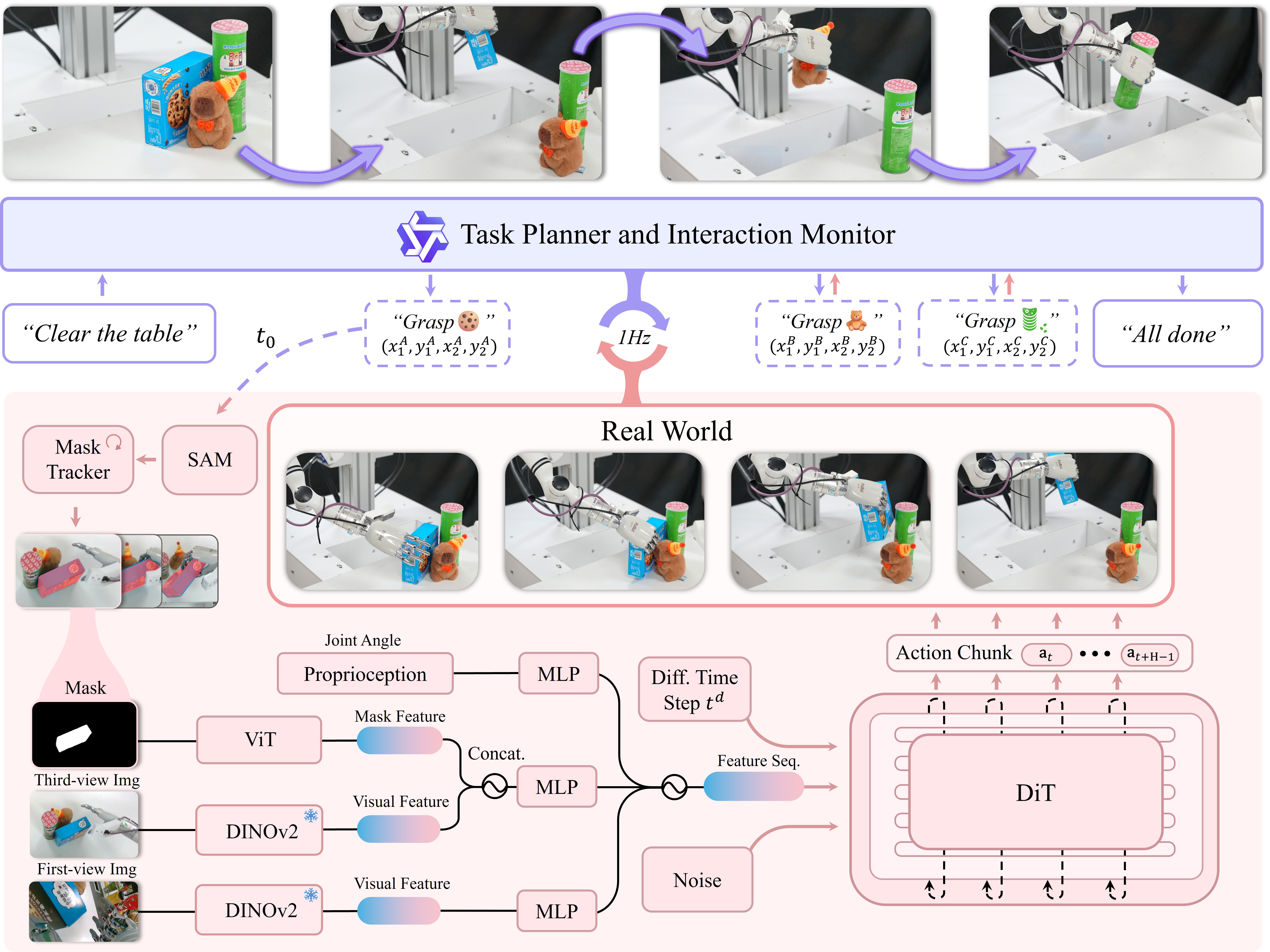}
    \caption{\textbf{Overview of DexGraspVLA}. A pre-trained VLM-based high-level planner (purple) decomposes prompts into object-level grasping instructions with bounding boxes. The diffusion-based low-level controller (pink) tracks the target mask, encodes multimodal observations (RGB images, mask, proprioception), and predicts an action chunk via a DiT model. The planner monitors execution and continually proposes new instructions based on the updated scene until the task is fully completed.}
    \label{fig:framework}
    \vspace{-10pt}
\end{figure*}

\paragraph{Dexterous Grasping.} Dexterous grasping methods are typically divided into two-stage and end-to-end approaches. Two-stage methods first generate a grasp pose—via sampling~\cite{zhang2024dexgraspnet,fang2025anydexgraspgeneraldexterousgrasping}, optimization~\cite{wang2023dexgraspnet,chen2024springgrasp}, or regression~\cite{li2022hgc,liu2020deep}—and reach it with motion planning. Though they benefit from modularity and synthetic data, their open-loop nature makes them vulnerable to disturbances and calibration errors. End-to-end methods learn grasping policies via reinforcement learning in massively parallel simulation~\cite{wan2023unidexgrasp++,zhang2024graspxl,singh2024dextrah}, which efficiently acquire emergent dexterity but suffer from sim-to-real gaps. In this work, we explore imitation learning from human demonstrations, which has shown promise on complex tasks~\cite{qin2022one, guzey2023see, lin2024learning}. Our core contribution is to address the central challenge of generalization in imitation learning~\cite{black2025pi}. We show that performing imitation learning on domain-invariant representations derived from foundation models enables strong generalization to unseen scenarios.

\paragraph{Foundation Models for Generalizable Robotic Policies.}
Vision and language foundation models pre-trained on web-scale data have shown impressive world knowledge and generalization~\cite{kirillov2023segment, oquab2023dinov2, qwen2.5-VL}, making them promising for robotics. A common approach, as seen in OpenVLA~\cite{kim2024openvla} and $\pi_0$~\cite{black2024pi0}, directly fine-tunes VLMs on robot data in the hope of transferring vision-language knowledge to the policy for broad generalization. However, this typically requires a massive amount of diverse demonstrations~\cite{o2023open}, yet still struggles with unseen scenarios and catastrophic forgetting. 
A more related line of work to ours instead leverages frozen foundation models to robustly infer task affordances—i.e., where and how to manipulate—in novel environments, guiding either motion planning~\cite{huang2024rekep,huang2023voxposer,pan2025omnimanip} or a learned action head~\cite{stoneopen}. However, the former often depends heavily on accurate calibration, involves considerable human design, or lacks robustness due to open-loop control. The latter still maps \emph{raw} visual inputs directly to actions, making it vulnerable to domain shift. In contrast, to achieve generalization across diverse real-world domains, our framework employs foundation models to iteratively transform free-form language prompts and diverse visual perceptions into domain-invariant representations. These representations enable imitation learning to be applied efficiently and effectively, collectively leading to robust generalization.

\section{Problem Formulation}
\label{sec:problem-setups}
Our goal is to develop a vision-based control policy for language-guided dexterous grasping, formulated as a sequential decision-making problem.
Initially, a language instruction $l$ is given, \emph{e.g.} ``\textit{grasp the toy}'', to directly specify the target object.
At each timestep $t$, the policy $\pi$ receives a first-view image $\mathbf{I}_{t}^{\mathrm{w}} \in \mathbb{R}^{H \times W \times 3}$ from the wrist camera ($H$ and $W$ denote the height and width of the image), a third-view image $\mathbf{I}_{t}^{\mathrm{h}} \in \mathbb{R}^{H \times W \times 3}$ from the head camera, and the robot proprioception $\mathbf{s}_t \in \mathbb{R}^{13}$ consisting of arm and hand joint angles $\mathbf{s}^{\mathrm{arm}}_t \in \mathbb{R}^{7}, \mathbf{s}^{\mathrm{hand}}_t \in \mathbb{R}^{6}$. 
Conditioned on these observations, the robot produces an action $\mathbf{a}_t = (\mathbf{a}^{\mathrm{arm}}_t, \mathbf{a}^{\mathrm{hand}}_t) \in \mathbb{R}^{13}$, where $ \mathbf{a}^{\mathrm{arm}}_t\in \mathbb{R}^{7}$ and $\mathbf{a}^{\mathrm{hand}}_t\in \mathbb{R}^{6}$ denote the target joint angles for arm and hand respectively, by sampling from the action distribution $\pi(\cdot | \{\mathbf{I}_j^{\mathrm{w}}\}_{j=0}^t, \{\mathbf{I}_j^{\mathrm{h}}\}_{j=0}^t, \{\mathbf{s}_j\}_{j=0}^t, l)$. 
This process continues until a termination condition is reached. The robot receives a binary reward $r \in \{0, 1\}$ indicating whether it has completed the instruction $l$ successfully. 
The goal of the policy $\pi$ is to maximize the expected reward $\mathbb{E}_{l,\{(\mathbf{I}^{\mathrm{w}}_{j}, \mathbf{I}^{\mathrm{h}}_{j}, \mathbf{s}_{j}, \mathbf{a}_{j})\}_{j=0}^{T}}[r].$

More generally, we consider cases where the user prompt $p$ may be a long-horizon task involving multiple grasping steps, such as ``\textit{clear the table}''. This requires the policy $\pi$ to reason about the prompt, decompose it into individual grasping instructions $\{l_i\}$, and complete them sequentially.

\section{Methods}
\label{sec:methods}

This section introduces DexGraspVLA, the first hierarchical VLA framework for dexterous grasping. We will first elaborate DexGraspVLA framework (\Cref{subsec:dexgraspvla-framework}) and then detail our data collection procedure (\Cref{subsec:data-collection}), which together enable the training of a dexterous grasping policy.

\subsection{DexGraspVLA Framework}
\label{subsec:dexgraspvla-framework}

As illustrated in \Cref{fig:framework}, DexGraspVLA adopts a hierarchical and modularized architecture composed of a planner and a controller. 
Below we explain how each part is designed.

\paragraph{Planner.}
We recognize that to achieve general dexterous grasping, the model must handle multimodal inputs, perform visual grounding, and conduct reasoning about user prompts. Building upon recent advances, we adopt an off-the-shelf pre-trained Qwen VLM~\cite{Qwen-VL, qwen2.5-VL} as a high-level planner to dynamically plan and monitor the dexterous grasping workflow. Given a user prompt $p$ (e.g., “clear the table”), the planner proposes a grasping instruction $l$ (e.g., “grasp the cookie”) as the first step.

For each $l$, the planner guides the low-level controller by marking the target object bounding box $(x_1, y_1, x_2, y_2)$ as task affordance in the head camera image $\mathbf{I}_{t_0}^{\mathrm{h}}$ at the initial timestep $t_0$. While the phrasing and content of language instruction can be diverse and flexible for different users and cases, \emph{i.e.}, showing \textit{domain-variance}, the bounding box is a consistent format for object localization regardless of the changes in language and visual inputs, \emph{i.e.}, achieving \textit{domain-invariance}. Thus, this transformation alleviates the learning challenge for the controller.

On issuing the bounding box, the planner monitors controller execution, resets robot after each grasp attempt, and proposes updated instruction $l$ until prompt $p$ is completed.

\paragraph{Controller.} 
Based on the bounding box \((x_1, y_1, x_2, y_2)\), the controller aims to grasp the intended object in cluttered environments. We feed this bounding box as input to SAM~\cite{kirillov2023segment} to obtain an initial binary mask \(\mathbf{m}_0 \in \{0, 1\}^{H \times W \times 1}\) of the target object and then use Cutie~\cite{cheng2024putting} to continuously track the mask over time, producing \(\mathbf{m}_t\) at each timestep \(t\). This ensures accurate identification in cluttered scenes throughout the process. The problem is to learn the policy $\pi$ that effectively models the action distribution $\pi(\cdot | \mathbf{I}_t^{\mathrm{w}}, \mathbf{I}_t^{\mathrm{h}}, \mathbf{s}_t, \mathbf{m}_t)$.

To achieve general-purpose dexterous grasping, the system must generalize effectively across diverse real-world scenarios. However, the high variability in raw visual inputs $\mathbf{I}_t^{\mathrm{w}}, \mathbf{I}_t^{\mathrm{h}}$ poses a fundamental challenge to learning task-critical representations. Traditional imitation learning approaches often fail catastrophically even under minor variations in objects or environmental conditions.
To address this issue, our solution is again to convert potentially \emph{domain-varying} inputs into \emph{domain-invariant} representations suitable for imitation learning. 
We recognize that \emph{while pixel-level perception vary widely, the fine-grained semantic features extracted by foundation models tend to be more robust and consistent}~\cite{tang2023emergentcorrespondenceimagediffusion,wang2023sparsedff}. 
Thus, we utilize a feature extractor \(\phi\), DINOv2~\cite{oquab2023dinov2}, to obtain features from raw images. At timestep \(t\),  we obtain head camera image features 
\(
\mathbf{z}^{\mathrm{h}}_t = \phi^{\mathrm{h}}(\mathbf{I}^{\mathrm{h}}_t) \in \mathbb{R}^{L^{\mathrm{h}} \times D^{\mathrm{h}}},
\)
and wrist camera image features
\(
\mathbf{z}^{\mathrm{w}}_t = \phi^{\mathrm{w}}(\mathbf{I}^{\mathrm{w}}_t) \in \mathbb{R}^{L^{\mathrm{w}} \times D^{\mathrm{w}}},
\)
where $L^{\mathrm{h}}, D^{\mathrm{h}}, L^{\mathrm{w}}, D^{\mathrm{w}}$ denote length and hidden dimension of the feature sequences for head and wrist respectively. As we show in \Cref{subsec:model-analysis}, these extracted features remain comparatively invariant to distracting visual factors.

Up to now, raw language and vision inputs, including instruction $l$ and images $\mathbf{I}_t^{\mathrm{w}}, \mathbf{I}_t^{\mathrm{h}}$, have been iteratively transformed into domain-invariant representations, including mask $\mathbf{m}_t$ and features $\mathbf{z}^{\mathrm{h}}_t, \mathbf{z}^{\mathrm{w}}_t$, by leveraging foundation models. This lays the stage for imitation learning. We now learn the policy $\pi$ that predicts an action chunk of horizon $H$ conditioning on these representations.

To fuse the object mask with head camera features, we project \(\mathbf{m}_t\) into the head image feature space using a randomly initialized ViT, producing \(\mathbf{z}^{\mathrm{m}}_t \in \mathbb{R}^{L^{\mathrm{h}} \times D^{\mathrm{h}}}\), and concatenate it with \(\mathbf{z}^{\mathrm{h}}_t\) patch-wise to obtain 
\(
\bar{\mathbf{z}}^{\mathrm{h}}_t \in \mathbb{R}^{L^{\mathrm{h}} \times 2D^{\mathrm{h}}}.
\)
Subsequently, we map \(\bar{\mathbf{z}}^{\mathrm{h}}_t\), wrist-camera features \(\mathbf{z}^{\mathrm{w}}_t\), and robot state \(\mathbf{s}_t\) into a common embedding space with separate MLPs, yielding \(\tilde{\mathbf{z}}^{\mathrm{h}}_t\), \(\tilde{\mathbf{z}}^{\mathrm{w}}_t\), and \(\tilde{\mathbf{z}}^{\mathrm{s}}_t\). These embeddings are then concatenated to form the full observation feature sequence 
\(
\tilde{\mathbf{z}}^{\mathrm{obs}}_t \in \mathbb{R}^{(1 + L^{\mathrm{h}} + L^{\mathrm{w}}) \times D}.
\)

For action prediction, we employ a DiT~\cite{peebles2023scalable} to generate multi-step actions, following the diffusion policy paradigm~\cite{chi2023diffusion, liu2024rdt}. At each timestep \(t\), we bundle the next \(H\) actions into a chunk \(\mathbf{A}_t = \mathbf{a}_{t:t+H} = [\mathbf{a}_t, \mathbf{a}_{t+1}, \dots, \mathbf{a}_{t+H-1}]\). During training, a random diffusion step \(t^d = k\) is sampled,
and Gaussian noise \(\boldsymbol{\epsilon}\) is added to \(\mathbf{A}_t\), yielding the noised action tokens \(\mathbf{x}_k = \alpha_k \mathbf{A}_t + \sigma_k \boldsymbol{\epsilon}, 
\)
where \(\alpha_k\) and \(\sigma_k\) are DDPM coefficients. We then feed \(\mathbf{x}_k\) into the DiT alongside the observation feature sequence \(\tilde{\mathbf{z}}^{\mathrm{obs}}_t\). Each DiT layer performs bidirectional self-attention over the action tokens, cross-attention to \(\tilde{\mathbf{z}}^{\mathrm{obs}}_t\), and MLP transformations, ultimately predicting the original noise \(\boldsymbol{\epsilon}\). By minimizing the noise prediction error, the model learns to reconstruct the ground-truth action chunk \(\mathbf{A}_t\). At inference time, iterative denoising steps recover the intended multi-step action sequence from the learned distribution, enabling imitation of multimodal behaviors. We also employ the receding horizon control strategy that only executes the first $H_a$ actions before generating a new action chunk prediction, enhancing responsiveness.

Overall, DexGraspVLA performs imitation learning on \emph{domain-invariant} representations derived from \emph{domain-varying} inputs via foundation models. This approach leverages the world knowledge and generalization capabilities of foundation models while effectively capturing the mapping from these abstracted representations to action output. 

\subsection{Data Collection}
\label{subsec:data-collection}

To train our dexterous grasping policy, we manually collect a dataset consisting of 2,094 successful demonstrations in cluttered scenes using 36 household objects varying in size, weight, geometry, texture, material, and category. Each episode
\(
  \tau = \{ (\mathbf{I}^{\mathrm{h}}_t, \mathbf{I}^{\mathrm{w}}_t, \mathbf{s}_t, \mathbf{m}_t, \mathbf{a}_t) \}_{t=0}^{T}
\)
records raw camera images \(\mathbf{I}^{\mathrm{h}}_t, \mathbf{I}^{\mathrm{w}}_t\), robot proprioception $\mathbf{s}_t$, object mask \(\mathbf{m}_t\), and action \(\mathbf{a}_t\) at each timestep \(t\). The mask $\mathbf{m}_t$ is labeled in the same way as in the controller. 
For each object, we place it randomly and collect multiple grasping demonstrations, with the surrounding objects randomized between episodes.
These demonstrations are performed at typical human motion speeds, taking about \SI{3.5}{\second} each. They undergo rigorous inspection to ensure quality. The DexGraspVLA controller is trained on this dataset with imitation learning.

\section{Experiments}
\label{sec:experiments}

In this section, we extensively evaluate DexGraspVLA. All experiments are conducted in a different environment from the demonstration setup, ensuring a "\emph{zero-shot}" setting to rigorously assess generalization to novel real-world scenarios. Our experiments seek to address the following questions: (1) \textbf{Large-scale Generalization} (\Cref{subsec:generalization-eval}): Can DexGraspVLA generalize to thousands of unseen object, lighting, and background combinations? (2) \textbf{Baseline Comparison} (\Cref{subsec:baseline-comparison}): How does its performance compare to baselines?  (3) \textbf{Ablation Study} (\Cref{subsec:baselines}): How much does imitation learning on domain-invariant representations improve generalization? (4) \textbf{Mechanism Analysis} (\Cref{subsec:model-analysis}): Are its internal model behaviors consistent under varying environments? (5) \textbf{Long-horizon Task} (\Cref{subsec:long-horizon-evaluation}): How effectively does DexGraspVLA handle free-form, long-horizon instructions? (6) \textbf{Extension to Nonprehensile Grasping} (\Cref{subsec:nonprehensile-grasping}): Can it be extended to other dexterous manipulation skills beyond grasping?

\subsection{Experiment Setups}
\label{subsec:experiment-setup}

\begin{figure}[t]
  \centering
  \includegraphics[width=0.7\linewidth]{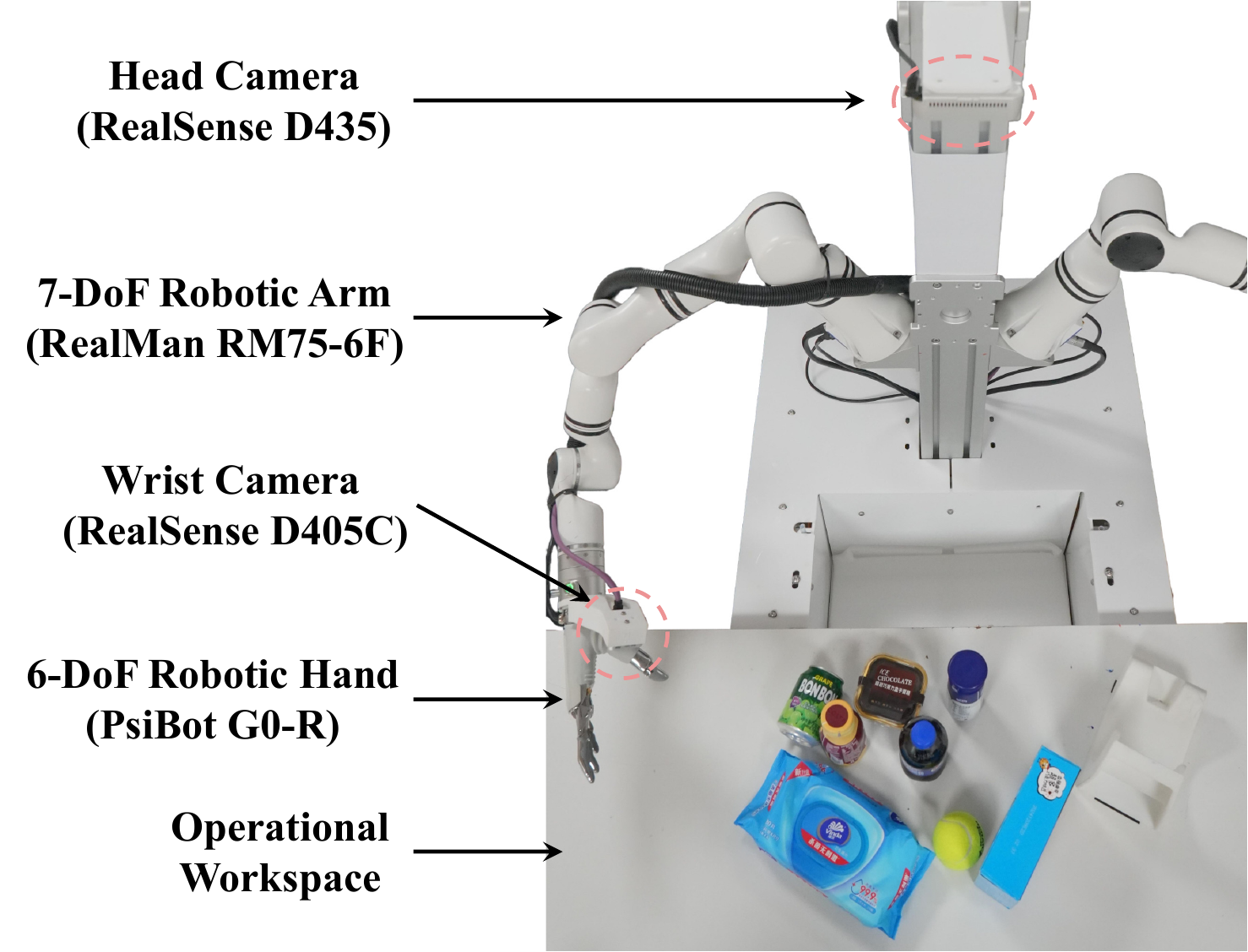}
  \caption{Our hardware platform.}
  \label{fig:robot-platform}
  \vspace{-10pt}
\end{figure}

\paragraph{Hardware Platform.} As shown in \Cref{fig:robot-platform}, our setup includes a 7-DoF RealMan RM75-6F arm and a 6-DoF PsiBot G0-R hand. A wrist-mounted RealSense D405C camera provides a first-person view, and a head-mounted D435 camera captures a third-person view. Objects are placed on a table in front, and the control frequency is 20 Hz.

\paragraph{Baselines.} 
We compare DexGraspVLA (Ours) with several state-of-the-art (SOTA) VLA baselines fine-tuned on our dataset, including full-parameter (Full FT) and LoRA fine-tuned variants of $\pi_0$~\cite{black2024pi0}, RDT~\cite{liu2024rdt}, OpenVLA~\cite{kim2024openvla}, and OpenVLA-OFT~\cite{kim2025fine}. We also evaluate two ablated versions of our method: 1) DINOv2-train: Identical to DexGraspVLA but with trainable DINOv2 encoders. 2) ViT-small: Identical to DexGraspVLA but replaces DINOv2 with smaller, trainable ViTs. Empirically, the ViT-small variant represents an enhanced version of Diffusion Policy~\cite{chi2023diffusion}, a SOTA imitation learning baseline. For all experiments, the high-level planner is based on Qwen-VL-Chat~\cite{Qwen-VL}, except in the long-horizon task (\Cref{subsec:long-horizon-evaluation}), where we use Qwen2.5-VL-72B-Instruct~\cite{qwen2.5-VL}. Implementation details are in \Cref{app:implementation-details}. 
To account for inference randomness, we report Ours@$k$ ($k=1,2,3$) in \Cref{subsec:generalization-eval}, where up to $k$ attempts are allowed per test. Ours@1 is equivalent to Ours. Re-grasps performed by the policy after an initial failure within a single attempt are allowed and not counted separately.

\begin{table*}[t]
    \centering
    \normalsize 

    \begin{subtable}[t]{0.45\textwidth}
        \centering
        \setlength{\tabcolsep}{0pt} 
        \renewcommand{\arraystretch}{1.1} 
        {\smaller
        \begin{tabular*}{\linewidth}{@{\extracolsep{\fill}}lccccc@{}}
            \toprule
            & \makecell{Seen\\Objects} &\makecell{Unseen\\Objects} & \makecell{Unseen\\Bgs.} & \makecell{Unseen\\Lights} & Aggr. \\
            \midrule
            OpenVLA (LoRA)      &33.3\%   &16.7\%   &14.6\%   &4.2\%  &12.9\%         \\
            OpenVLA-OFT (LoRA)  &25.0\%   &29.2\%   &31.3\%   &31.3\% &30.3\%   \\
            RDT (Full FT)       & 25.0\%  & 25.0\%  &31.3\%   &35.4\% &31.1\%   \\
            $\pi_0$ (LoRA)      & 58.3\%  & 45.8\%  &14.6\%   &10.4\% &22.7\%  \\
            $\pi_0$ (Full FT)   & 75.0\%  & 45.8\%  &20.8\%   &20.8\% &30.3\%  \\
            Ours                & \textbf{91.7}\% & \textbf{91.7}\% & \textbf{89.6}\% &\textbf{93.8}\%   &\textbf{91.7}\%     \\
            \bottomrule
        \end{tabular*}
        }
        \caption{Dexterous grasping in cluttered scenes.}
        \label{tab:baselines}
    \end{subtable}
    \hfill
    \begin{subtable}[t]{0.49\textwidth}
        \centering
        \setlength{\tabcolsep}{0pt} 
        \renewcommand{\arraystretch}{1.1} 
        {\smaller
        \begin{tabular*}{\linewidth}{@{\extracolsep{\fill}}lccccc@{}}
            \toprule
            & \makecell{Clear\\Table} & \makecell{Grasp\\Green} & \makecell{Grasp\\Bottles}
            & \makecell{Grasp\\Food} & \makecell{Aggr.} \\
            \midrule
            Task Success Rate                   & 95.8\%  & 87.5\% & 91.7\% & 83.3\% & 89.6\% \\
            Avg. Attempts per Grasp             & 1.09    & 1.14   & 1.09   & 1.19   & 1.12   \\
            Planner: Instruction Proposal      & 100.0\% & 92.6\% & 94.3\% & 88.1\% & 94.3\% \\
            Planner: BBox Accuracy              & 98.7\%  & 98.2\% & 98.1\% & 98.3\% & 98.4\% \\
            Controller: Grasping                & 91.0\%  & 92.6\% & 92.5\% & 91.5\% & 92.2\% \\
            Planner: Completion Check          & 98.7\%  & 94.4\% & 96.2\% & 94.9\% & 96.3\% \\
            \bottomrule
        \end{tabular*}
        }
        \caption{Long-horizon task performance of DexGraspVLA.}
        \label{tab:long-horizon-prompts}
    \end{subtable}

    \vspace{1em} 
    \begin{subtable}[t]{0.30\textwidth}
        \centering
        \setlength{\tabcolsep}{0pt}
        \renewcommand{\arraystretch}{1.1}
        {\smaller
        \begin{tabular*}{\linewidth}{@{\extracolsep{\fill}}lcccc@{}}
            \toprule
            & \makecell{Unseen\\Objects} & \makecell{Unseen\\Bgs.} & \makecell{Unseen\\Lights} & Aggr. \\
            \midrule
            Ours@1 & 91.1\% & 90.5\% & 90.9\% & 90.8\% \\
            Ours@2 & 95.3\% & 94.2\% & 95.1\% & 94.7\% \\
            Ours@3 & 96.7\% & 96.7\% & 97.4\% & 96.9\% \\
            \bottomrule
        \end{tabular*}
        }
        \caption{Large-scale generalization evaluation of DexGraspVLA on dexterous grasping.}
        \label{tab:large-scale-generalization}
    \end{subtable}
    \hfill
    \begin{subtable}[t]{0.25\textwidth}
        \centering
        \setlength{\tabcolsep}{0pt}
        \renewcommand{\arraystretch}{1.1}
        {\smaller
        \begin{tabular*}{\linewidth}{@{\extracolsep{\fill}}lccc@{}}
            \toprule
            & \makecell{Seen\\Objects} & \makecell{Unseen\\Objects} & Aggr. \\
            \midrule
            ViT-small    & 60.0\% & 35.0\% & 50.5\% \\
            DINOv2-train & 30.0\% & 43.5\% & 34.8\% \\
            Ours         & \textbf{98.5}\% & \textbf{98.8}\% & \textbf{98.6}\% \\
            \bottomrule
        \end{tabular*}
        }
        \caption{Ablation results on single-object grasping.}
        \label{tab:grasp-single-object}
    \end{subtable}
    \hfill
    \begin{subtable}[t]{0.32\textwidth}
        \centering
        \setlength{\tabcolsep}{0pt}
        \renewcommand{\arraystretch}{1.1}
        {\smaller
        \begin{tabular*}{\linewidth}{@{\extracolsep{\fill}}lcccc@{}}
            \toprule
            & \makecell{Unseen\\Objects} & \makecell{Unseen\\Bgs.} & \makecell{Unseen\\Lights} & Aggr.\\
            \midrule
            ViT-small    & 61.1\% & 37.5\% & 22.2\% & 39.6\% \\
            DINOv2-train & 66.7\% & 70.8\% & 55.6\% & 66.0\% \\
            Ours         & \textbf{88.9}\% & \textbf{86.1}\% & \textbf{77.8}\% & \textbf{84.7}\% \\
            \bottomrule
        \end{tabular*}
        }
        \caption{Nonprehensile grasping performance.}
        \label{tab:nonprehensile-grasping}
    \end{subtable}

    \caption{Comprehensive evaluation of DexGraspVLA and baselines across tasks. Bgs.: Backgrounds; Aggr.: Aggregated.}
    \vspace{-10pt}
    \label{tab:all-results}
\end{table*}
\subsection{Large-Scale Generalization Evaluation}
\label{subsec:generalization-eval}

\paragraph{Tasks.} We curate 360 unseen objects, 6 unseen backgrounds, and 3 unseen lighting conditions.
The objects span diverse sizes, weights, geometries, textures, materials, and categories, while remaining graspable by our dexterous hand. Backgrounds and lighting conditions are selected to be visually distinct.
We evaluate generalization through three grasping tasks in cluttered scenes (around 6 objects per scene): (1) \emph{Unseen Objects}: Each of the 360 objects is grasped once in a random scene on a white table under white light (360 tests). (2) \emph{Unseen Backgrounds}: A subset of 103 objects $\mathcal{S}$ forms 103 scenes per background under white light, totaling 618 tests. (3) \emph{Unseen Lightings}: The same $\mathcal{S}$ forms 103 scenes per lighting condition on a white table (309 tests). Details can be found in \Cref{app:experiment-details}.

\paragraph{Metric.} A grasp is successful if the object is held \SI{10}{\centi\metre} above the table for \SI{20}{\second}. Success rate is the ratio of successes to total tests; aggregated performance is a weighted average by task proportion.

\paragraph{Results.} We present the quantitative results in \Cref{tab:large-scale-generalization}. From the 1st row (“Ours@1”), DexGraspVLA achieves a 91.1\% single-attempt success rate on 360 unseen objects, 90.5\% on 6 unseen backgrounds, and 90.9\% under 3 unseen lighting conditions, yielding a 90.8\% aggregated success rate. 
These results demonstrate robust and accurate control of the dexterous hand to grasp specified objects from clutter in diverse unseen conditions, without domain-specific fine-tuning. This highlights strong generalization and suggests that our framework substantially alleviates the overfitting challenge in imitation learning. We further analyze the source of this generalization in \Cref{subsec:model-analysis} and extend its application in \Cref{subsec:nonprehensile-grasping}.

Qualitatively, DexGraspVLA robustly handles challenging cases involving transparent, deformable, reflective, or background-camouflaged objects. It also dexterously adapts to diverse geometries and poses --- e.g., grasping a bottle from the side, picking up a small earbud case from the top, or retrieving an awkwardly placed box. The closed-loop policy enables re-grasping after failed attempts and tolerates human-induced perturbations by tracking object motion. Such robustness stems from three factors: first, foundation-model-based perception ensures semantic consistency under appearance variation; second, imitation learning avoids the need for explicit object modeling; and third, diffusion-based action head captures multi-modal action distributions.

From the 2nd and 3rd rows (“Ours@2” and “Ours@3”), we observe that allowing up to three attempts further boosts performance to 96.9\%, indicating the capacity to reach even higher success rates.
Finally, our model takes around \SI{6}{\second} to grasp an object, which is close to that of humans and ensures practical usability in real-world scenarios.

\subsection{Baseline Comparison}
\label{subsec:baseline-comparison}

\paragraph{Tasks \& Metrics.} We adopt the same setup as \Cref{subsec:generalization-eval} but on a smaller scale for baseline comparison. The tasks involve 24 unseen objects, 2 unseen backgrounds, and 2 unseen lighting conditions. We also include 12 seen objects under white background and lighting (\emph{Seen Objects}). Metrics remain unchanged; details are in \Cref{app:experiment-details}.

\paragraph{Results.} As shown in \Cref{tab:baselines}, DexGraspVLA consistently achieves 90+\% success across all settings, significantly outperforming fine-tuned VLA models. While $\pi_0$ (Full FT) reaches 75\% on seen objects, its performance drops sharply under visual variations. Similar declines are observed for $\pi_0$ (LoRA) and OpenVLA (LoRA), suggesting overfitting to training language and visual domains. Notably, RDT also uses frozen vision and language foundation models like ours and shows more consistent performance, but still falls short. This suggests that bounding boxes offer stronger grounding than language encoding, and that DINOv2 better preserves visual details than SigLIP~\cite{zhai2023sigmoid}. Overall, these results validate the design of DexGraspVLA and its superior generalization performance.

\subsection{Ablation Study}
\label{subsec:baselines}

\paragraph{Tasks \& Metrics.} To compare DexGraspVLA with ablated variants that learn directly from raw visual inputs without frozen vision encoders, we conduct single-object grasping experiments using 13 seen and 8 unseen objects. Each object is tested at five table locations with two trials per location, yielding 210 tests under white tabletop and lighting. Success rates are computed as in \Cref{subsec:generalization-eval}.

\paragraph{Results.} \Cref{tab:grasp-single-object} shows that DexGraspVLA (Ours) consistently achieves over 98\% success on both seen and unseen objects, significantly outperforming DINOv2-train and ViT-small variants. Its near-perfect performance in a zero-shot setting indicates strong robustness to domain shift. Interestingly, performance on unseen objects slightly exceeds that on seen ones, suggesting that the model learns the grasping task itself rather than overfitting to training data. In contrast, baselines that map raw inputs to actions fail to generalize, as perceptual changes easily push them out of distribution.

\subsection{Internal Model Behavior Analysis}
\label{subsec:model-analysis}

To further validate our design, we examine whether internal model behavior remains consistent under varying visual conditions, as shown in \Cref{fig:dino-attn}. We test DexGraspVLA on the same cluttered scene (9 objects, target: “grasp the blue yogurt in the middle”) across four environments: a white table, a calibration board, a colorful tablecloth, and the same tablecloth under disco lighting. For clarity, we display only the tabletop region; full images are in \Cref{app:experiment-details}. While the head images (1st row) appear to be markedly diverse, the DINOv2 features (2nd row) look rather consistent. These features are visualized by mapping principal components to RGB channels as done in \citet{oquab2023dinov2}. Across environments, the object properties are robustly maintained and matched, which fundamentally allows DexGraspVLA trained on a single data domain to generalize. The third row shows that Cutie accurately tracks the object, providing the correct guidance to the controller. Based on the domain-invariant mask and the DINOv2 features, the DiT action head now predicts the subsequent actions. In the fourth row, we average and normalize all cross-attentions to the head image from DiT. We find that all attention maps exhibit the same behavior of focusing on the target object instead of being distracted by environments. The fifth row overlays the attention map on the raw image to confirm the reasonable attention pattern. All visualization details are provided in \Cref{app:experiment-details}. Therefore, we substantiate that DexGraspVLA indeed transforms perceptually diverse raw inputs into invariant representations, on which it effectively applies imitation learning to model the data distribution, explaining its superior generalization performance. 

\subsection{Long-Horizon Task Evaluation}
\label{subsec:long-horizon-evaluation}

\paragraph{Tasks.}

We evaluate DexGraspVLA on long-horizon grasping tasks. We use four types of prompts—“\textit{Clear the table}”, “\textit{Grasp all bottles}”, “\textit{Grasp all green objects}”, and “\textit{Grasp all food}”—which require commonsense and physical reasoning to identify targets sequentially. Each prompt is evaluated in 24 randomly configured cluttered scenes. “\textit{Clear the table}” scenes include three unseen objects; others involve 3–4 unseen objects, with two being relevant. 

\paragraph{Metric.}
For each task, we report the task success rate as the proportion of tests that fully complete all required stages. We further report the average grasping attempts per object in the successful tests, along with success rates for instruction proposal, bounding box prediction, completion check of the planner, and grasp execution of the controller.

\paragraph{Results.} \Cref{tab:long-horizon-prompts} shows that DexGraspVLA achieves an 89.6\% aggregated task success rate across four long-horizon prompts, with each target object attempted slightly more than once. The high-level planner grounds prompt semantics on the observation and proposes correct instructions with a 94.3\% average success rate. Its bounding box prediction accuracy is consistently above 98\%, which we further substantiate with evaluations in distraction conditions in \Cref{app:additional-experiments}. The low-level controller, leveraging its robust and generalizable grasping policy, executes individual grasps with over 91\% success, enabling reliable multi-step completion. Additionally, the planner detects task completion with over 94\% accuracy, preventing redundant actions. These results highlight the synergy between the high-level and low-level modules in DexGraspVLA, showcasing the effectiveness of its hierarchical framework for long-horizon tasks. An example can be found in \Cref{app:additional-results}.

\subsection{Extension to Nonprehensile Grasping}
\label{subsec:nonprehensile-grasping}

\begin{figure}[t]
  \centering
  \includegraphics[width=\linewidth]{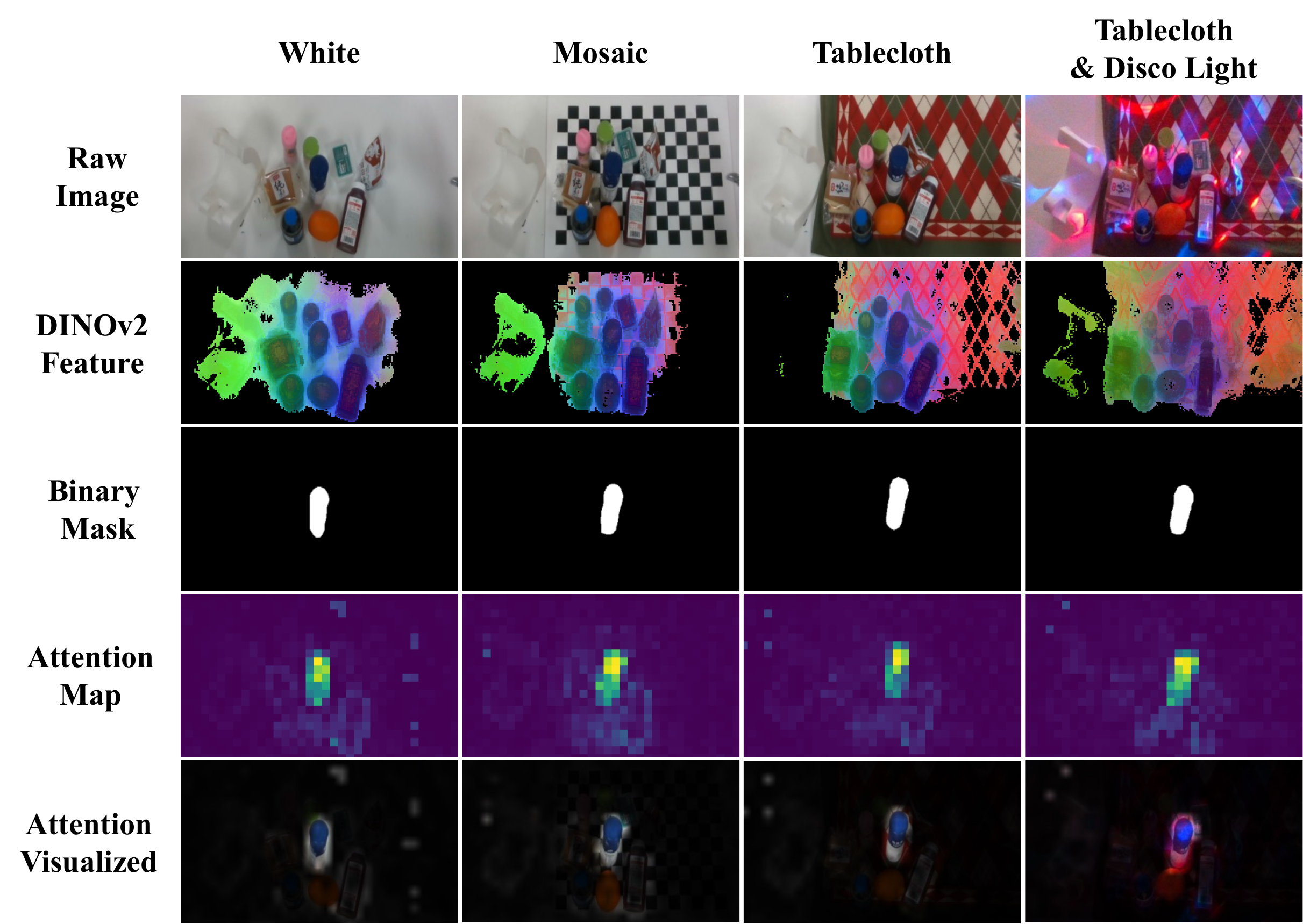}
  \caption{\textbf{DexGraspVLA is robust to environmental variations.} The same cluttered scene (1st row) is arranged in four visually different environments (four columns). DINOv2 features (2nd row), masks (3rd row), and attention maps (4th row) are consistent across variations. The 5th row confirms DexGraspVLA is attending to the correct object.}
  \label{fig:dino-attn}
  \vspace{-10pt}
\end{figure}

\paragraph{Tasks \& Metric.} To show applicability beyond dexterous grasping, we apply DexGraspVLA to a nonprehensile grasping task (\Cref{fig:teaser} last row). We curate 32 flat, wide-surface objects (e.g., plates, boxes, and books) that are difficult to grasp directly and collect 1,029 human demonstrations in cluttered scenes. In these demos, the robot first performs a pre-grasp manipulation by pushing the object toward the table edge, creating an accessible pose, and then executes a final grasp. 
We keep the DexGraspVLA planner unchanged and train the controller on this dataset; details are provided in Appendix~\ref{app:implementation-details}. To evaluate generalization, we curate 18 unseen nonprehensile objects and design three types of tasks: (1) \emph{Unseen Objects} (36 tests): Each object is placed in two cluttered scenes with varying poses on a white table under white light. (2) \emph{Unseen Lighting} (36 tests): Same protocol under disco light. (3) \emph{Unseen Backgrounds} (72 tests): Same protocol on a wooden tabletop or a yellow tablecloth.
 Success rates are reported as in \Cref{subsec:generalization-eval}.

\paragraph{Results.} As shown in \Cref{tab:nonprehensile-grasping}, DexGraspVLA achieves an aggregated generalization performance of 84.7\% in the nonprehensile grasping task, showing strong robustness to unseen object appearances, shapes, physical properties, as well as novel backgrounds and lightings—significantly outperforming ablated variants. We observe that DexGraspVLA reliably adapts to object poses, pushing until it extends sufficiently over the edge, followed by a stable grasp. This task is particularly challenging for parallel-jaw grippers, highlighting the dexterity we exhibit. Moreover, DexGraspVLA seamlessly extends to this new task without architectural changes, reflecting three key aspects of generality: (1) the high-level planner’s grounding and reasoning ability; (2) the use of bounding boxes as affordance guidance; and (3) applying imitation learning on domain-invariant representations iteratively obtained from foundation models.

\section{Limitation and Conclusion}
\label{sec:conclusion}

This paper presents DexGraspVLA, a hierarchical VLA framework aiming for robust generalization in language-guided dexterous grasping and beyond. By leveraging a pre-trained VLM as the high-level planner and vision foundation models in the low-level controller, the system transforms multimodal inputs into domain-invariant representations and learns robust closed-loop policies via imitation learning. Our large-scale evaluations show over 90\% grasping success across thousands of unseen cluttered scenes in a zero-shot setting, with empirical evidence of consistent internal behavior. DexGraspVLA also handles free-form long-horizon prompts, recovers from failures, and extends to nonprehensile grasping, demonstrating broad applicability. While effective, it does not yet address functional grasping and subsequent manipulation, nor does it incorporate tactile sensing. In future work, we aim to extend the high-level planner to generate more fine-grained affordance and learn a task-oriented manipulation controller that also integrates tactile feedback, further broadening the scope of DexGraspVLA.

\bibliography{aaai2026}

@article{steiner2022train,
  title={How to train your ViT? Data, Augmentation, and Regularization in Vision Transformers},
  author={Steiner, Andreas Peter and Kolesnikov, Alexander and Zhai, Xiaohua and Wightman, Ross and Uszkoreit, Jakob and Beyer, Lucas},
  journal={Transactions on Machine Learning Research},
  year={2022}
}

@inproceedings{wang2023dexgraspnet,
  title={Dexgraspnet: A large-scale robotic dexterous grasp dataset for general objects based on simulation},
  author={Wang, Ruicheng and Zhang, Jialiang and Chen, Jiayi and Xu, Yinzhen and Li, Puhao and Liu, Tengyu and Wang, He},
  booktitle={2023 IEEE International Conference on Robotics and Automation (ICRA)},
  pages={11359--11366},
  year={2023},
  organization={IEEE}
}

@inproceedings{zhang2024dexgraspnet,
  title={DexGraspNet 2.0: Learning Generative Dexterous Grasping in Large-scale Synthetic Cluttered Scenes},
  author={Zhang, Jialiang and Liu, Haoran and Li, Danshi and Yu, XinQiang and Geng, Haoran and Ding, Yufei and Chen, Jiayi and Wang, He},
  booktitle={8th Annual Conference on Robot Learning},
  year={2024}
}

@inproceedings{zhang2024graspxl,
  title={Graspxl: Generating grasping motions for diverse objects at scale},
  author={Zhang, Hui and Christen, Sammy and Fan, Zicong and Hilliges, Otmar and Song, Jie},
  booktitle={European Conference on Computer Vision},
  pages={386--403},
  year={2024},
  organization={Springer}
}

@inproceedings{kirillov2023segment,
  title={Segment anything},
  author={Kirillov, Alexander and Mintun, Eric and Ravi, Nikhila and Mao, Hanzi and Rolland, Chloe and Gustafson, Laura and Xiao, Tete and Whitehead, Spencer and Berg, Alexander C and Lo, Wan-Yen and others},
  booktitle={Proceedings of the IEEE/CVF International Conference on Computer Vision},
  pages={4015--4026},
  year={2023}
}

@inproceedings{cheng2024putting,
  title={Putting the object back into video object segmentation},
  author={Cheng, Ho Kei and Oh, Seoung Wug and Price, Brian and Lee, Joon-Young and Schwing, Alexander},
  booktitle={Proceedings of the IEEE/CVF Conference on Computer Vision and Pattern Recognition},
  pages={3151--3161},
  year={2024}
}

@article{oquab2023dinov2,
  title={Dinov2: Learning robust visual features without supervision},
  author={Oquab, Maxime and Darcet, Timoth{\'e}e and Moutakanni, Th{\'e}o and Vo, Huy and Szafraniec, Marc and Khalidov, Vasil and Fernandez, Pierre and Haziza, Daniel and Massa, Francisco and El-Nouby, Alaaeldin and others},
  journal={Transactions on Machine Learning Research},
  year={2023}
}

@inproceedings{wan2023unidexgrasp++,
  title={Unidexgrasp++: Improving dexterous grasping policy learning via geometry-aware curriculum and iterative generalist-specialist learning},
  author={Wan, Weikang and Geng, Haoran and Liu, Yun and Shan, Zikang and Yang, Yaodong and Yi, Li and Wang, He},
  booktitle={Proceedings of the IEEE/CVF International Conference on Computer Vision},
  pages={3891--3902},
  year={2023}
}

@article{o2023open,
  title={Open x-embodiment: Robotic learning datasets and rt-x models},
  author={O'Neill, Abby and Rehman, Abdul and Gupta, Abhinav and Maddukuri, Abhiram and Gupta, Abhishek and Padalkar, Abhishek and Lee, Abraham and Pooley, Acorn and Gupta, Agrim and Mandlekar, Ajay and others},
  journal={arXiv preprint arXiv:2310.08864},
  year={2023}
}

@article{liu2024rdt,
  title={Rdt-1b: a diffusion foundation model for bimanual manipulation},
  author={Liu, Songming and Wu, Lingxuan and Li, Bangguo and Tan, Hengkai and Chen, Huayu and Wang, Zhengyi and Xu, Ke and Su, Hang and Zhu, Jun},
  journal={arXiv preprint arXiv:2410.07864},
  year={2024}
}

@article{black2024pi0,
    title={$\pi_0$: {A} {V}ision-{L}anguage-{A}ction {F}low {M}odel for {G}eneral {R}obot {C}ontrol},
    author={Black, Kevin and Brown, Noah and Driess, Danny and Esmail, Adnan and Equi, Michael and Finn, Chelsea and Fusai, Niccolo and Groom, Lachy and Hausman, Karol and Ichter, Brian and others},
    journal={arXiv preprint arXiv:2410.24164},
    year={2024}
}

@article{kim2024openvla,
  title={OpenVLA: An Open-Source Vision-Language-Action Model},
  author={Kim, Moo Jin and Pertsch, Karl and Karamcheti, Siddharth and Xiao, Ted and Balakrishna, Ashwin and Nair, Suraj and Rafailov, Rafael and Foster, Ethan and Lam, Grace and Sanketi, Pannag and others},
  journal={arXiv preprint arXiv:2406.09246},
  year={2024}
}

@article{akkaya2019solving,
  title={Solving rubik's cube with a robot hand},
  author={Akkaya, Ilge and Andrychowicz, Marcin and Chociej, Maciek and Litwin, Mateusz and McGrew, Bob and Petron, Arthur and Paino, Alex and Plappert, Matthias and Powell, Glenn and Ribas, Raphael and others},
  journal={arXiv preprint arXiv:1910.07113},
  year={2019}
}

@inproceedings{handa2022dextreme,
  title={DeXtreme: Transfer of Agile In-hand Manipulation from Simulation to Reality},
  author={Handa, Ankur and Allshire, Arthur and Makoviychuk, Viktor and Petrenko, Aleksei and Singh, Ritvik and Liu, Jingzhou and Makoviichuk, Denys and Van Wyk, Karl and Zhurkevich, Alexander and Sundaralingam, Balakumar and others},
  booktitle={2023 IEEE International Conference on Robotics and Automation (ICRA)},
  pages={5977--5984},
  year={2023}
}

@article{yang2024anyrotate,
  title={AnyRotate: Gravity-Invariant In-Hand Object Rotation with Sim-to-Real Touch},
  author={Yang, Max and Lu, Chenghua and Church, Alex and Lin, Yijiong and Ford, Chris and Li, Haoran and Psomopoulou, Efi and Barton, David AW and Lepora, Nathan F},
  journal={arXiv preprint arXiv:2405.07391},
  year={2024}
}

@inproceedings{pitz2023dextrous,
  title={Dextrous tactile in-hand manipulation using a modular reinforcement learning architecture},
  author={Pitz, Johannes and R{\"o}stel, Lennart and Sievers, Leon and B{\"a}uml, Berthold},
  booktitle={2023 IEEE International Conference on Robotics and Automation (ICRA)},
  pages={1852--1858},
  year={2023},
  organization={IEEE}
}

@article{zakka2023robopianist,
  title={RoboPianist: A Benchmark for High-Dimensional Robot Control},
  author={Zakka, Kevin and Smith, Laura and Gileadi, Nimrod and Howell, Taylor and Peng, Xue Bin and Singh, Sumeet and Tassa, Yuval and Florence, Pete and Zeng, Andy and Abbeel, Pieter},
  journal={arXiv preprint arXiv:2304.04150},
  year={2023}
}

@article{chen2022towards,
  title={Towards human-level bimanual dexterous manipulation with reinforcement learning},
  author={Chen, Yuanpei and Wu, Tianhao and Wang, Shengjie and Feng, Xidong and Jiang, Jiechuan and Lu, Zongqing and McAleer, Stephen and Dong, Hao and Zhu, Song-Chun and Yang, Yaodong},
  journal={Advances in Neural Information Processing Systems},
  volume={35},
  pages={5150--5163},
  year={2022}
}

@inproceedings{guzey2023see,
  title={See to touch: Learning tactile dexterity through visual incentives},
  author={Guzey, Irmak and Dai, Yinlong and Evans, Ben and Chintala, Soumith and Pinto, Lerrel},
  booktitle={2024 IEEE International Conference on Robotics and Automation (ICRA)},
  pages={13825--13832},
  year={2024},
  organization={IEEE}
}

@inproceedings{huang2023dynamic,
  title={Dynamic handover: Throw and catch with bimanual hands},
  author={Huang, Binghao and Chen, Yuanpei and Wang, Tianyu and Qin, Yuzhe and Yang, Yaodong and Atanasov, Nikolay and Wang, Xiaolong},
  booktitle={7th Annual Conference on Robot Learning},
  year={2023}
}

@inproceedings{huang2023voxposer,
  title={Voxposer: Composable 3d value maps for robotic manipulation with language models},
  author={Huang, Wenlong and Wang, Chen and Zhang, Ruohan and Li, Yunzhu and Wu, Jiajun and Fei-Fei, Li},
  booktitle={Conference on Robot Learning},
  pages={540--562},
  year={2023}
}

@article{qin2022one,
  title={From one hand to multiple hands: Imitation learning for dexterous manipulation from single-camera teleoperation},
  author={Qin, Yuzhe and Su, Hao and Wang, Xiaolong},
  journal={IEEE Robotics and Automation Letters},
  volume={7},
  number={4},
  pages={10873--10881},
  year={2022},
  publisher={IEEE}
}

@article{chen2024springgrasp,
  title={SpringGrasp: An optimization pipeline for robust and compliant dexterous pre-grasp synthesis},
  author={Chen, Sirui and Bohg, Jeannette and Liu, C Karen},
  journal={arXiv preprint arXiv:2404.13532},
  year={2024}
}

@inproceedings{chen2023sequential,
  title={Sequential dexterity: Chaining dexterous policies for long-horizon manipulation},
  author={Chen, Yuanpei and Wang, Chen and Fei-Fei, Li and Liu, C Karen},
  booktitle={Conference on Robot Learning},
  pages={3809--3829},
  year={2023}
}

@article{lin2024learning,
  title={Learning Visuotactile Skills with Two Multifingered Hands},
  author={Lin, Toru and Zhang, Yu and Li, Qiyang and Qi, Haozhi and Yi, Brent and Levine, Sergey and Malik, Jitendra},
  journal={arXiv preprint arXiv:2404.16823},
  year={2024}
}

@article{lin2024twisting,
  title={Twisting lids off with two hands},
  author={Lin, Toru and Yin, Zhao-Heng and Qi, Haozhi and Abbeel, Pieter and Malik, Jitendra},
  journal={arXiv preprint arXiv:2403.02338},
  year={2024}
}

@inproceedings{qi2023general,
  title={General in-hand object rotation with vision and touch},
  author={Qi, Haozhi and Yi, Brent and Suresh, Sudharshan and Lambeta, Mike and Ma, Yi and Calandra, Roberto and Malik, Jitendra},
  booktitle={Conference on Robot Learning},
  pages={2549--2564},
  year={2023},
  organization={PMLR}
}

@article{Qwen-VL,
  title={Qwen-VL: A Frontier Large Vision-Language Model with Versatile Abilities},
  author={Bai, Jinze and Bai, Shuai and Yang, Shusheng and Wang, Shijie and Tan, Sinan and Wang, Peng and Lin, Junyang and Zhou, Chang and Zhou, Jingren},
  journal={arXiv preprint arXiv:2308.12966},
  year={2023}
}

@inproceedings{huang2024rekep,
  title={Rekep: Spatio-temporal reasoning of relational keypoint constraints for robotic manipulation},
  author={Huang, Wenlong and Wang, Chen and Li, Yunzhu and Zhang, Ruohan and Fei-Fei, Li},
  booktitle={8th Annual Conference on Robot Learning},
  year={2024}
}

@article{chi2023diffusion,
  title={Diffusion policy: Visuomotor policy learning via action diffusion},
  author={Chi, Cheng and Xu, Zhenjia and Feng, Siyuan and Cousineau, Eric and Du, Yilun and Burchfiel, Benjamin and Tedrake, Russ and Song, Shuran},
  journal={The International Journal of Robotics Research},
  year={2023},
  publisher={SAGE Publications Sage UK: London, England}
}

@inproceedings{turpin2022grasp,
  title={Grasp’d: Differentiable contact-rich grasp synthesis for multi-fingered hands},
  author={Turpin, Dylan and Wang, Liquan and Heiden, Eric and Chen, Yun-Chun and Macklin, Miles and Tsogkas, Stavros and Dickinson, Sven and Garg, Animesh},
  booktitle={European Conference on Computer Vision},
  pages={201--221},
  year={2022},
  organization={Springer}
}

@inproceedings{turpin2023fast,
  title={Fast-Grasp'D: Dexterous Multi-finger Grasp Generation Through Differentiable Simulation},
  author={Turpin, Dylan and Zhong, Tao and Zhang, Shutong and Zhu, Guanglei and Heiden, Eric and Macklin, Miles and Tsogkas, Stavros and Dickinson, Sven and Garg, Animesh},
  booktitle={2023 IEEE International Conference on Robotics and Automation (ICRA)},
  pages={8082--8089},
  year={2023},
  organization={IEEE}
}

@inproceedings{li2022hgc,
  title={HGC-Net: Deep anthropomorphic hand grasping in clutter},
  author={Li, Yiming and Wei, Wei and Li, Daheng and Wang, Peng and Li, Wanyi and Zhong, Jun},
  booktitle={2022 International Conference on Robotics and Automation (ICRA)},
  pages={714--720},
  year={2022},
  organization={IEEE}
}

@inproceedings{liu2020deep,
  title={Deep differentiable grasp planner for high-dof grippers},
  author={Liu, Min and Pan, Zherong and Xu, Kai and Ganguly, Kanishka and Manocha, Dinesh},
  booktitle={Robotics: Science and Systems},
  year={2020}
}

@article{singh2024dextrah,
  title={DextrAH-RGB: Visuomotor Policies to Grasp Anything with Dexterous Hands},
  author={Singh, Ritvik and Allshire, Arthur and Handa, Ankur and Ratliff, Nathan and Van Wyk, Karl},
  journal={arXiv preprint arXiv:2412.01791},
  year={2024}
}

@inproceedings{wang2023sparsedff,
  title={Sparsedff: Sparse-view feature distillation for one-shot dexterous manipulation},
  author={Wang, Qianxu and Zhang, Haotong and Deng, Congyue and You, Yang and Dong, Hao and Zhu, Yixin and Guibas, Leonidas},
  booktitle={The Twelfth International Conference on Learning Representations},
  year={2023}
}

@inproceedings{peebles2023scalable,
  title={Scalable diffusion models with transformers},
  author={Peebles, William and Xie, Saining},
  booktitle={Proceedings of the IEEE/CVF International Conference on Computer Vision},
  pages={4195--4205},
  year={2023}
}

@article{li2024immiscible,
  title={Immiscible diffusion: Accelerating diffusion training with noise assignment},
  author={Li, Yiheng and Jiang, Heyang and Kodaira, Akio and Tomizuka, Masayoshi and Keutzer, Kurt and Xu, Chenfeng},
  journal={arXiv preprint arXiv:2406.12303},
  year={2024}
}

@inproceedings{songdenoising,
  title={Denoising Diffusion Implicit Models},
  author={Song, Jiaming and Meng, Chenlin and Ermon, Stefano},
  booktitle={International Conference on Learning Representations},
  year={2020}
}

@inproceedings{radford2021learning,
  title={Learning transferable visual models from natural language supervision},
  author={Radford, Alec and Kim, Jong Wook and Hallacy, Chris and Ramesh, Aditya and Goh, Gabriel and Agarwal, Sandhini and Sastry, Girish and Askell, Amanda and Mishkin, Pamela and Clark, Jack and others},
  booktitle={International conference on machine learning},
  pages={8748--8763},
  year={2021},
  organization={PMLR}
}

@article{hurst2024gpt,
  title={Gpt-4o system card},
  author={Hurst, Aaron and Lerer, Adam and Goucher, Adam P and Perelman, Adam and Ramesh, Aditya and Clark, Aidan and Ostrow, AJ and Welihinda, Akila and Hayes, Alan and Radford, Alec and others},
  journal={arXiv preprint arXiv:2410.21276},
  year={2024}
}

@misc{qwen2.5-VL,
    title = {Qwen2.5-VL},
    url = {https://qwenlm.github.io/blog/qwen2.5-vl/},
    author = {Qwen Team},
    month = {January},
    year = {2025}
}

@inproceedings{
black2025pi,
title={\${\textbackslash}pi\_\{0.5\}\$: a Vision-Language-Action Model with Open-World Generalization},
author={Physical Intelligence},
booktitle={9th Annual Conference on Robot Learning},
year={2025},
url={https://openreview.net/forum?id=vlhoswksBO}
}

@inproceedings{zhou2023learning,
  title={Learning to grasp the ungraspable with emergent extrinsic dexterity},
  author={Zhou, Wenxuan and Held, David},
  booktitle={Conference on Robot Learning},
  pages={150--160},
  year={2023},
  organization={PMLR}
}

@inproceedings{kwon2023efficient,
  title={Efficient Memory Management for Large Language Model Serving with PagedAttention},
  author={Woosuk Kwon and Zhuohan Li and Siyuan Zhuang and Ying Sheng and Lianmin Zheng and Cody Hao Yu and Joseph E. Gonzalez and Hao Zhang and Ion Stoica},
  booktitle={Proceedings of the ACM SIGOPS 29th Symposium on Operating Systems Principles},
  year={2023}
}

@article{zhong2025survey,
  title={A Survey on Vision-Language-Action Models: An Action Tokenization Perspective},
  author={Zhong, Yifan and Bai, Fengshuo and Cai, Shaofei and Huang, Xuchuan and Chen, Zhang and Zhang, Xiaowei and Wang, Yuanfei and Guo, Shaoyang and Guan, Tianrui and Lui, Ka Nam and others},
  journal={arXiv preprint arXiv:2507.01925},
  year={2025}
}

@inproceedings{pan2025omnimanip,
  title={Omnimanip: Towards general robotic manipulation via object-centric interaction primitives as spatial constraints},
  author={Pan, Mingjie and Zhang, Jiyao and Wu, Tianshu and Zhao, Yinghao and Gao, Wenlong and Dong, Hao},
  booktitle={Proceedings of the Computer Vision and Pattern Recognition Conference},
  pages={17359--17369},
  year={2025}
}

@inproceedings{stoneopen,
  title={Open-World Object Manipulation using Pre-Trained Vision-Language Models},
  author={Stone, Austin and Xiao, Ted and Lu, Yao and Gopalakrishnan, Keerthana and Lee, Kuang-Huei and Vuong, Quan and Wohlhart, Paul and Kirmani, Sean and Zitkovich, Brianna and Xia, Fei and others},
  booktitle={7th Annual Conference on Robot Learning},
  year={2023}
}

@misc{fang2025anydexgraspgeneraldexterousgrasping,
      title={AnyDexGrasp: General Dexterous Grasping for Different Hands with Human-level Learning Efficiency}, 
      author={Hao-Shu Fang and Hengxu Yan and Zhenyu Tang and Hongjie Fang and Chenxi Wang and Cewu Lu},
      year={2025},
      eprint={2502.16420},
      archivePrefix={arXiv},
      primaryClass={cs.RO},
      url={https://arxiv.org/abs/2502.16420}, 
}

@article{kim2025fine,
  title={Fine-tuning vision-language-action models: Optimizing speed and success},
  author={Kim, Moo Jin and Finn, Chelsea and Liang, Percy},
  journal={arXiv preprint arXiv:2502.19645},
  year={2025}
}

@inproceedings{zhai2023sigmoid,
  title={Sigmoid loss for language image pre-training},
  author={Zhai, Xiaohua and Mustafa, Basil and Kolesnikov, Alexander and Beyer, Lucas},
  booktitle={Proceedings of the IEEE/CVF international conference on computer vision},
  pages={11975--11986},
  year={2023}
}

@misc{tang2023emergentcorrespondenceimagediffusion,
      title={Emergent Correspondence from Image Diffusion}, 
      author={Luming Tang and Menglin Jia and Qianqian Wang and Cheng Perng Phoo and Bharath Hariharan},
      year={2023},
      eprint={2306.03881},
      archivePrefix={arXiv},
      primaryClass={cs.CV},
      url={https://arxiv.org/abs/2306.03881}, 
}

@misc{gemini25,
  title        = {Gemini 2.5: Our most intelligent AI model},
  author       = {Gemini team},
  year         = 2025,
  url          = {https://blog.google/technology/google-deepmind/gemini-model-thinking-updates-march-2025}
}

@misc{openai2024gpt4ocard,
  title        = {GPT-4o System Card},
  author       = {OpenAI},
  year         = 2024,
  url          = {https://arxiv.org/abs/2410.21276},
  eprint       = {2410.21276},
  archiveprefix = {arXiv},
  primaryclass = {cs.CL}
}

@article{team2024gemma,
  title={Gemma: Open models based on gemini research and technology},
  author={Team, Gemma and Mesnard, Thomas and Hardin, Cassidy and Dadashi, Robert and Bhupatiraju, Surya and Pathak, Shreya and Sifre, Laurent and Rivi{\`e}re, Morgane and Kale, Mihir Sanjay and Love, Juliette and others},
  journal={arXiv preprint arXiv:2403.08295},
  year={2024}
}

\clearpage
\appendix
\renewcommand{\thesubsection}{\thesection.\arabic{subsection}}
\captionsetup[table]{justification=justified}
\section{Implementation Details}
\label{app:implementation-details}

In this section, we present the details of DexGraspVLA implementation (\Cref{app:dexgraspvla-details}), baseline implementation (\Cref{app:baseline-details}), and dataset collection (\Cref{app:data-details}).

\subsection{Details of DexGraspVLA Implementation}
\label{app:dexgraspvla-details}

\paragraph{Planner.}
The high-level planner operates as described in \Cref{subsec:dexgraspvla-framework}. By leveraging an off-the-shelf VLM as the planner, our framework gains remarkable flexibility, enabling easy utilization of more advanced models for enhanced performance. Our observations indicate that Qwen2.5-VL-72B-Instruct~\cite{qwen2.5-VL} outperforms Qwen-VL-Chat~\cite{Qwen-VL} in reasoning and instruction following, leading to improved long-horizon task completion. Therefore, we base the DexGraspVLA planner on Qwen2.5-VL-72B-Instruct in the long-horizon tasks and provide our prompts below. 

These prompts mainly instruct the VLM to function as DexGraspVLA planner via four sub-tasks, including (1) Instruction Proposal: proposing the current grasping instruction $l$ based on the user prompt $p$, (2) Bounding Box Prediction: marking the target object bounding box, (3) Grasp Outcome Verification: checking if the grasp has succeeded, and (4) Prompt Completion Check: evaluating whether the entire user prompt is fully fulfilled.
Since instruction proposal, bounding box prediction, and prompt completion check only require information within the operational workspace on the table, we crop the relevant region from the head camera image and fill the remaining area with white pixels. The resulting cropped image is used as the planner's visual input for these sub-tasks.

To start with, when a user prompt $p$ is provided, the planner first determines which object in the scene should be grasped next. This step involves interpreting the prompt in context and selecting the best matching object from the current visual input.

\begin{tcolorbox}[breakable]

You are controlling a robotic arm that needs to complete the following user prompt: <user\_prompt>.

I will show you two images. 
The initial image (before any actions) is: <initial\_head\_image>. 
The current image (after the latest action) is: <current\_head\_image>.

Your task is to select the \textbf{best object to grasp next} from the current image. 

To identify objects, \textbf{use common sense and everyday knowledge} to infer what each item is.

For example, recognize cups, bottles, fruits, snacks, boxes, tools, etc.
\newline\newline
When choosing the best object to grasp, follow these principles:

1. Prefer objects on the right, then center, then left.

2. Avoid objects that are blocked or surrounded.

3. Avoid grasping objects that would cause other items to topple.

4. Select objects that best match the user prompt.
\newline\newline
Please output ONLY ONE object that the robot should grasp next.
\newline\newline
Return format (in English, natural language):

A short sentence precisely describing the target object, including:

- color.

- shape.

- relative position (e.g., "on the right", "in front", "next to the red box").
\newline\newline
Example:

Grasp the blue cube on the right side of the table.

\end{tcolorbox}

After deciding on the next object to grasp, the planner proceeds to locate this object in the image by predicting its bounding box using the following prompts. The generated grasping instruction is used as input to this localization module.

\begin{tcolorbox}[breakable]

You are a robotic vision assistant. Your task is to locate the object described below in the given image: <current\_head\_image> and return its bounding box.
\newline\newline
Grasping instruction: <grasping\_instruction>.
\newline\newline
Instructions:

1. Carefully read the grasping instruction and match the target object to the best-fitting visible object in the image.

2. Select EXACTLY ONE object that best matches the description.

3. For the selected object, return the following in strict JSON format:

- "bbox\_2d": [x1, y1, x2, y2] (integer pixel coordinates, top-left to bottom-right)

- "label": a short 2-4 word name, (e.g. "blue cup")

- "description": a complete, natural-language description of the object's appearance and position
\newline\newline
Requirements:

- Only return one object.

- Coordinates must be valid and within image boundaries.

- Do not guess if the object is not visible.

\end{tcolorbox}

During the controller’s execution, the planner verifies whether the object has been successfully grasped, using the following prompt.

\begin{tcolorbox}[breakable]

I will show you two images. 
The top-down view from the head camera is: <current\_head\_image>.
The close-up view from the wrist camera is: <current\_wrist\_image>.
\newline\newline
Grasping instruction: <grasping\_instruction>.
\newline\newline
Task:

Determine whether the robotic arm has \textbf{successfully grasped the target object}.
\newline\newline
You should consider:

- Whether the target object is still visible on the table.

- Whether the object is securely held in the robotic hand.
\newline\newline
Output format: 

A reasoning and a boolean value (True=successfully grasped, False=not grasped).
\newline\newline
Keep it short and simple.

\end{tcolorbox}

Upon a successful grasp, it triggers a scripted placing motion. After each grasp attempt, the planner resets the robot to the initial state and checks whether the user prompt has been fulfilled with the following prompt.

\begin{tcolorbox}[breakable]

The robot is trying to complete the following user prompt: <user\_prompt>.

I will show you two images. 
The initial image (before any actions) is: <initial\_head\_image>. 
The current image (after the latest action) is: <current\_head\_image>.

Please compare the two images and determine whether the user prompt has been fully completed.
\newline\newline
Instructions:

- Only consider visible 3D objects.

- If all target objects have been removed or grasped, return True.

- If some relevant objects remain, return False.
\newline\newline
Output format: 

A reasoning and a boolean value (True=completed, False=not completed).
\newline\newline
Example:

All blue objects have been removed from the table: True.

\end{tcolorbox}

In our experiments, we either query the online APIs of these models or host them on an 8-A800 GPU server by ourselves with vLLM~\cite{kwon2023efficient}. When hosting Qwen2.5-VL-72B-Instruct, we employ Qwen2.5-VL-7B-Instruct for speculative decoding to accelerate inference.

\paragraph{Controller.} We first elaborate on the implementation details for the controller in the general dexterous grasping experiments. All raw images are produced by head and wrist cameras at a resolution of $640\times480\times3$. Correspondingly, the resolution of mask is $640\times480\times1$. 
Through preliminary model selection, we decide to use DINOv2 ViT-B/14 as the feature extractor $\phi^{\mathrm{h}}$ for head camera images and DINOv2 ViT-L/14 as the feature extractor $\phi^{\mathrm{w}}$ for wrist camera images. Before feeding images into DINOv2, we resize them to $518\times518\times3$. During training, we apply domain randomization via color jittering. Finally, the images are normalized and fed into DINOv2 models. This leads to features $\mathbf{z}^{\mathrm{h}}_t \in\mathbb{R}^{1369\times768}$ and $\mathbf{z}^{\mathrm{w}}_t \in\mathbb{R}^{1369\times1024}$. By processing the mask $\mathbf{m}_t$ with a randomly initialized ViT, we extract its features $\mathbf{z}^{\mathrm{m}}_t \in\mathbb{R}^{1369\times768}$. Patch-wise concatenation of $\mathbf{z}^{\mathrm{h}}_t$ and $\mathbf{z}^{\mathrm{m}}_t$ leads to $\bar{\mathbf{z}}^{\mathrm{h}}_t\in\mathbb{R}^{1369\times1536}$. We then project $\bar{\mathbf{z}}^{\mathrm{h}}_t, \mathbf{z}^{\mathrm{w}}_t, \mathbf{s}_t$ to the same feature space of dimension $1024$ with separate MLPs, yielding $\tilde{\mathbf{z}}^{\mathrm{h}}_t\in\mathbb{R}^{1369\times1024}, \tilde{\mathbf{z}}^{\mathrm{w}}_t\in\mathbb{R}^{1369\times1024}, \tilde{\mathbf{z}}^{\mathrm{s}}_t\in\mathbb{R}^{1\times1024}$, and concatenate them to form the full observation feature sequence $\tilde{\mathbf{z}}^{\mathrm{obs}}_t = (\tilde{\mathbf{z}}^{\mathrm{h}}_t, \tilde{\mathbf{z}}^{\mathrm{w}}_t, \tilde{\mathbf{z}}^{\mathrm{s}}_t)\in\mathbb{R}^{2739\times1024}$.

For action modeling, we define an action chunk horizon of $H = 64$. When we add noise to the action during training, we employ Immiscible Diffusion~\cite{li2024immiscible} to improve data-noise mapping. The noised action chunk $\hat{\mathbf{A}}_t$ belongs to $\mathbb{R}^{64\times13}$.

The DiT implementation is based on the original DiT paper~\cite{peebles2023scalable}, diffusion policy~\cite{chi2023diffusion}, and RDT~\cite{liu2024rdt}. It first embeds the diffusion timestep to the same hidden space as $\tilde{\mathbf{z}}^{\mathrm{obs}}_t$, yielding $\tilde{\mathbf{z}}^{\mathrm{d}}_t\in\mathbb{R}^{1\times1024}$, and concatenates it with $\tilde{\mathbf{z}}^{\mathrm{obs}}_t$ to form the condition sequence $\tilde{\mathbf{z}}_t = (\tilde{\mathbf{z}}^{\mathrm{obs}}_t, \tilde{\mathbf{z}}^{\mathrm{d}}_t) \in\mathbb{R}^{2740\times1024}$. We project the noised action chunk to the same hidden space, deriving $\tilde{\mathbf{z}}^{\mathrm{A}}_t\in\mathbb{R}^{64\times1024}$, and feed it into DiT. Each DiT layer performs bi-directional attention within action tokens, cross-attention to the condition sequence, and MLP projections. Finally, the output is projected back to the action space to be the model's prediction of noise. During training, we compute MSE loss between the noise prediction and ground truth, and back-propagate the gradient to update all trainable parameters. During inference, we start from Gaussian noise and iteratively denoise it using DDIM sampling~\cite{songdenoising}. At each step, the DiT predicts the noise given the condition sequence, and we update the action chunk using the DDIM scheduler until we obtain the final action. The controller executes the first six actions in the predicted action chunk before making a new prediction.

In total, the controller possesses 163M trainable parameters. To accelerate training, we utilize bfloat16 mixed-precision training, reducing memory usage and improving computational efficiency. Additionally, we employ FusedAdamW as the optimizer to further speed up training through optimized memory access and fused kernel execution. With these techniques, we train the controller for 84 epochs over our dataset on an 8-A800 GPU server, which takes less than one day to complete. 
All hyper-parameters in our implementation are presented in \Cref{tab:hyperparam}.

In the nonprehensile grasping experiments, we keep most of the hyper-parameters the same but make the following changes: we use DINOv2 ViT-B/14 as the feature extractors $\phi^{\mathrm{h}}, \phi^{\mathrm{w}}$ for both head and wrist camera images, and the action horizon is set to $100$. This controller has 106M trainable parameters and is trained for 200 epochs on an 8-A800 GPU server, which takes approximately two days to finish.

\begin{figure*}[t]
    \centering

    \begin{subfigure}[t]{0.48\linewidth}
        \centering
        \includegraphics[width=\linewidth]{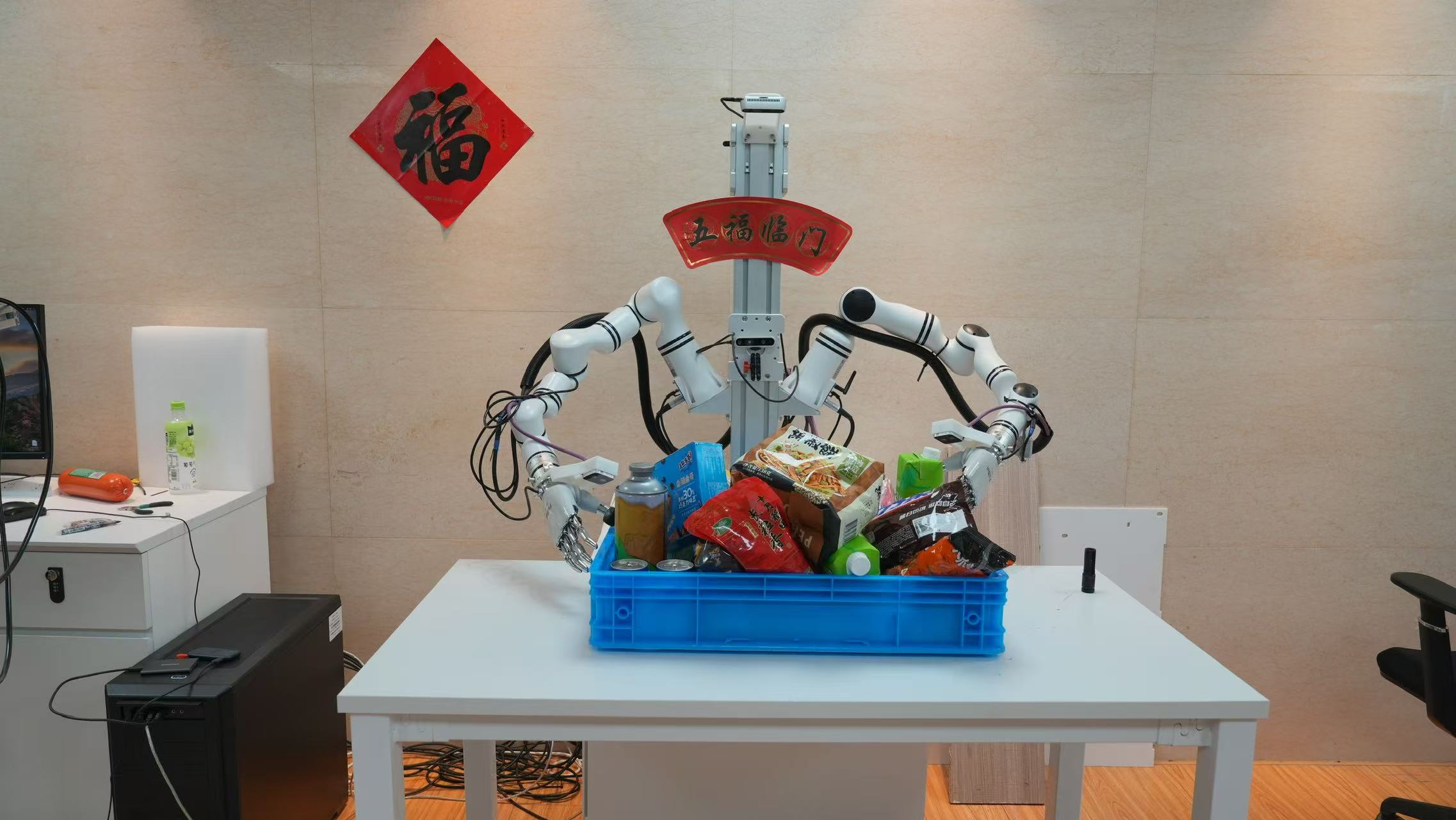}
        \captionsetup{justification=centering}
        \caption{Our data collection site.}
        \label{fig:data-collection-env}
    \end{subfigure}
    \hfill
    \begin{subfigure}[t]{0.48\linewidth}
        \centering
        \includegraphics[width=\linewidth]{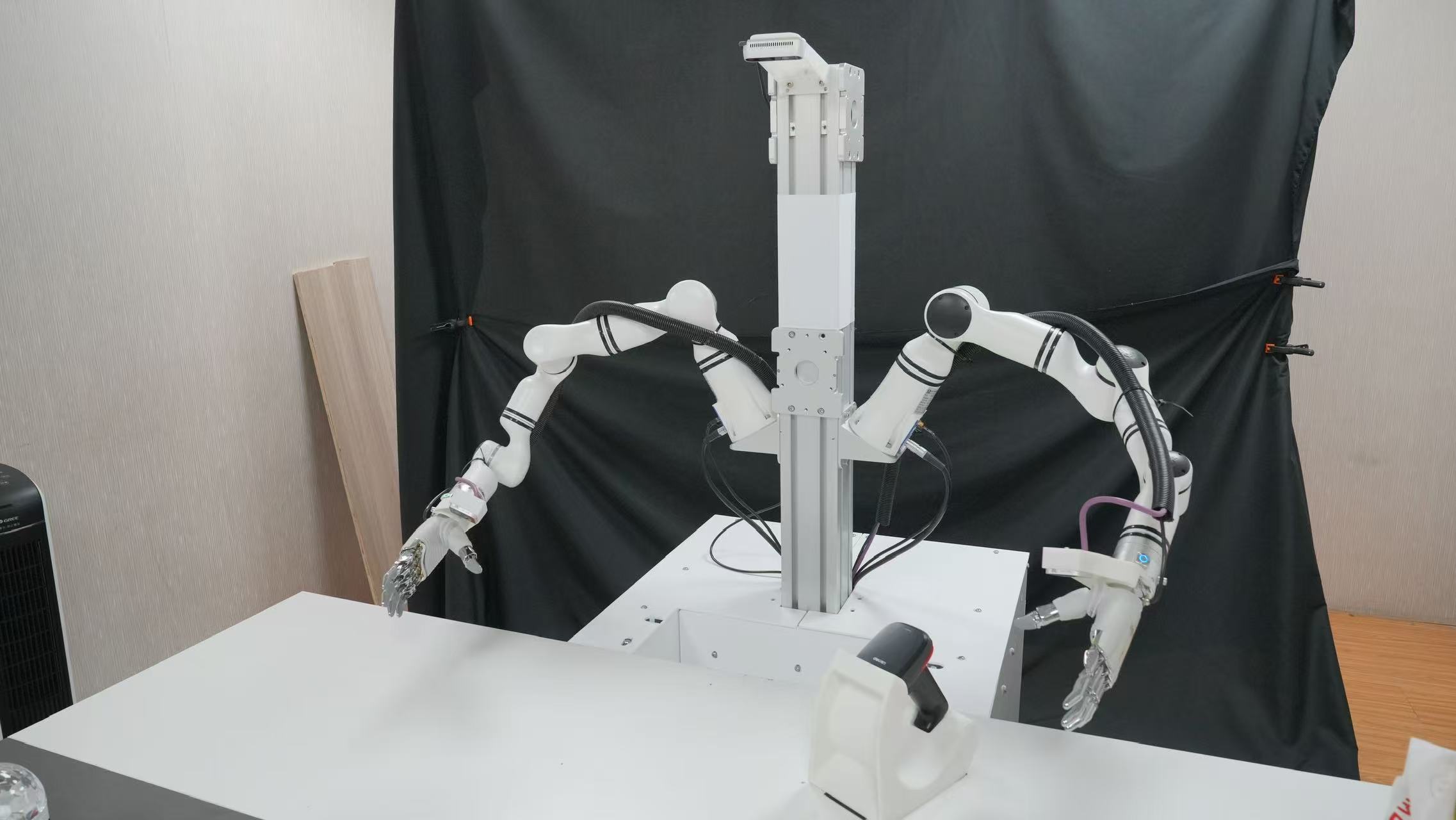}
        \captionsetup{justification=centering}
        \caption{The test environment for experiments.}
        \label{fig:test-env}
    \end{subfigure}

    \caption{A comparison of the data collection and test environments, located in different rooms. The visual scenes captured by the robot’s cameras differ significantly, especially for the wrist camera.}
    \label{fig:zero-shot-env}
\end{figure*}

\begin{table}[t]
\centering
\normalsize
\begin{tabular}{l|l}
\toprule
Hyper-parameter            & Value               \\ \midrule
epoch                      & 84                  \\
learning rate              & 0.0001              \\
learning rate scheduler    & cosine              \\
learning rate warmup steps & 2000                \\
weight decay               & 0.0001              \\
AdamW betas                & {[}0.95, 0.999{]}   \\
seed                       & 42                  \\
batch size per GPU         & 48                  \\
action horizon             & 64                  \\
number of DiT layers       & 12                  \\
number of DiT heads        & 8                   \\
attention dropout          & 0.1                 \\
noise scheduler            & DDIMScheduler       \\
num\_train\_timesteps      & 50                  \\
beta\_start                & 0.0001              \\
beta\_end                  & 0.02                \\
beta\_schedule             & squaredcos\_cap\_v2 \\
clip\_sample               & True                \\
set\_alpha\_to\_one        & True                \\
steps\_offset              & 0                   \\
prediction\_type           & epsilon             \\
num\_inference\_steps      & 16                  \\ \bottomrule
\end{tabular}
\caption{Hyper-parameters of DexGraspVLA.}
\label{tab:hyperparam}
\end{table}

\subsection{Details of Baseline Implementation}
\label{app:baseline-details}

\paragraph{Baselines.}
In the general dexterous grasping experiments, we fine-tune several state-of-the-art VLA models on our datasets following their official instructions.

Since our datasets do not contain language annotations, we first construct language instructions for each episode by manually annotating the target object. We then use LLMs, including GPT-4o~\cite{openai2024gpt4ocard} and Gemini 2.5 Pro~\cite{gemini25}, to expand and diversify the instructions. All generated instructions are manually verified.

For $\pi_0$, we perform both full-parameter and LoRA fine-tuning on 8 A800 GPUs. In the LoRA setup, we use a LoRA rank of 16 for the Gemma~\cite{team2024gemma} backbone and 32 for the action expert. The action horizon is 50, with a total batch size of 256, and the model is fine-tuned for 30K steps.

For RDT, we perform full-parameter fine-tuning on 8 A800 GPUs. The action horizon is 64, with a total batch size of 256, and training runs for 200K steps.

For OpenVLA, we perform LoRA fine-tuning on 4 A800 GPUs, using a total batch size of 16, LoRA rank of 32, and 60K fine-tuning steps. Note that OpenVLA does not support action chunking.

For OpenVLA-OFT, we apply LoRA fine-tuning on 8 A800 GPUs, with a total batch size of 32, LoRA rank of 8, action horizon of 25, and a total of 30K fine-tuning steps.

\paragraph{Ablation.}
In both general dexterous grasping and nonprehensile grasping experiments, DexGraspVLA (DINOv2-train) is the same as DexGraspVLA (Ours) described in \Cref{app:dexgraspvla-details} except that the two DINOv2 models are trainable instead of frozen. DexGraspVLA (ViT-small) is the same as DexGraspVLA (Ours) except that the two DINOv2 models are replaced with two small trainable pre-trained ViTs (the R26-S-32 ResNet-ViT hybrid from \citet{steiner2022train}). Correspondingly, we resize the images to $224\times224\times3$ to feed them into ViT-small. Each image is split into 49 patches, and the feature dimension is 384.

\subsection{Details of Data Collection}
\label{app:data-details}

\begin{figure*}[t]
  \centering
  \includegraphics[width=0.85\linewidth]{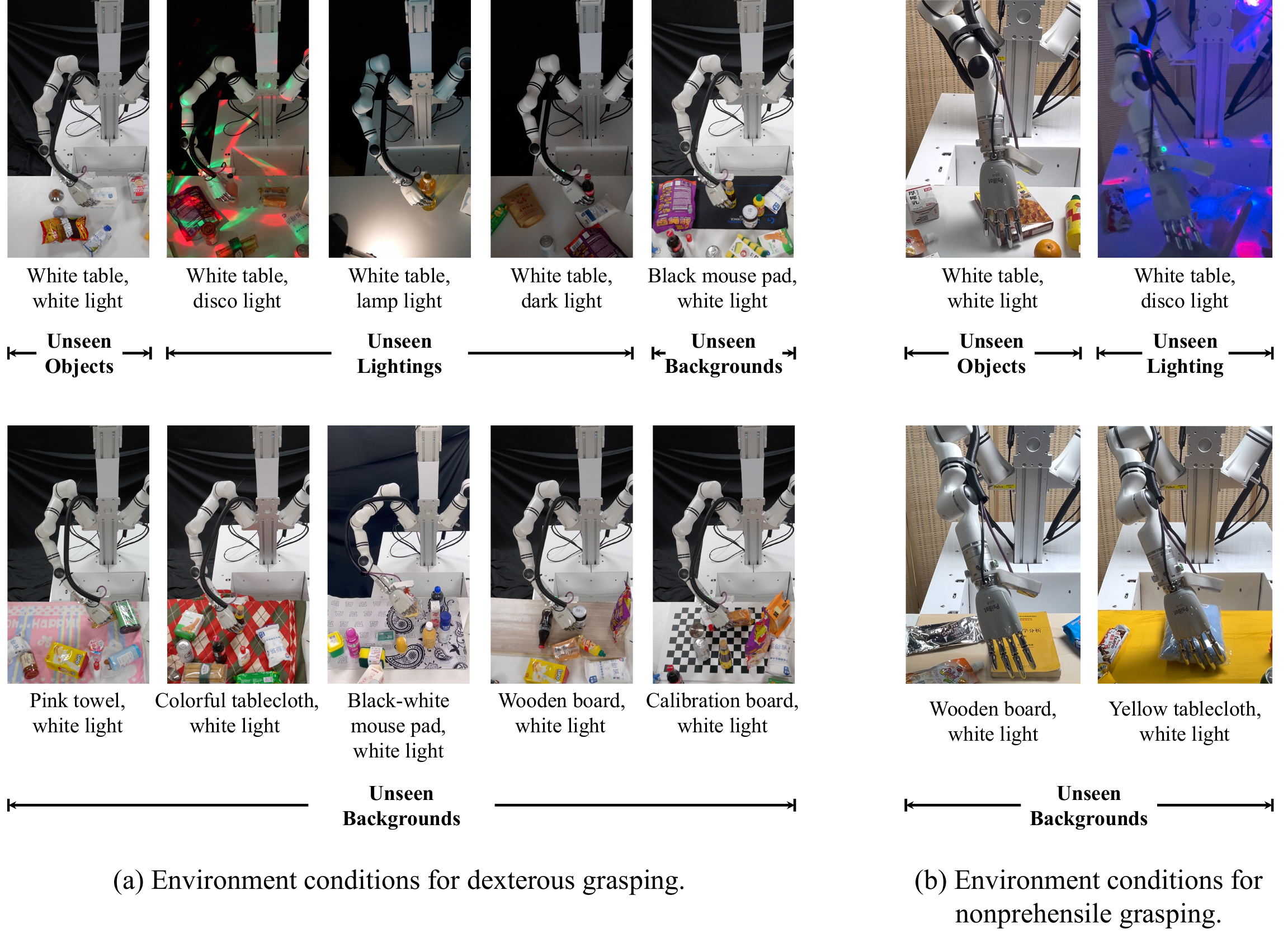}
  \caption{Environment conditions used in our generalization evaluations of dexterous grasping (\Cref{subsec:generalization-eval}) and nonprehensile grasping (\Cref{subsec:nonprehensile-grasping}).}
  \label{fig:generalization-eval-env}
  \vspace{-10pt}
\end{figure*}

\begin{figure*}[t]
  \centering
  \includegraphics[width=0.8\textwidth]{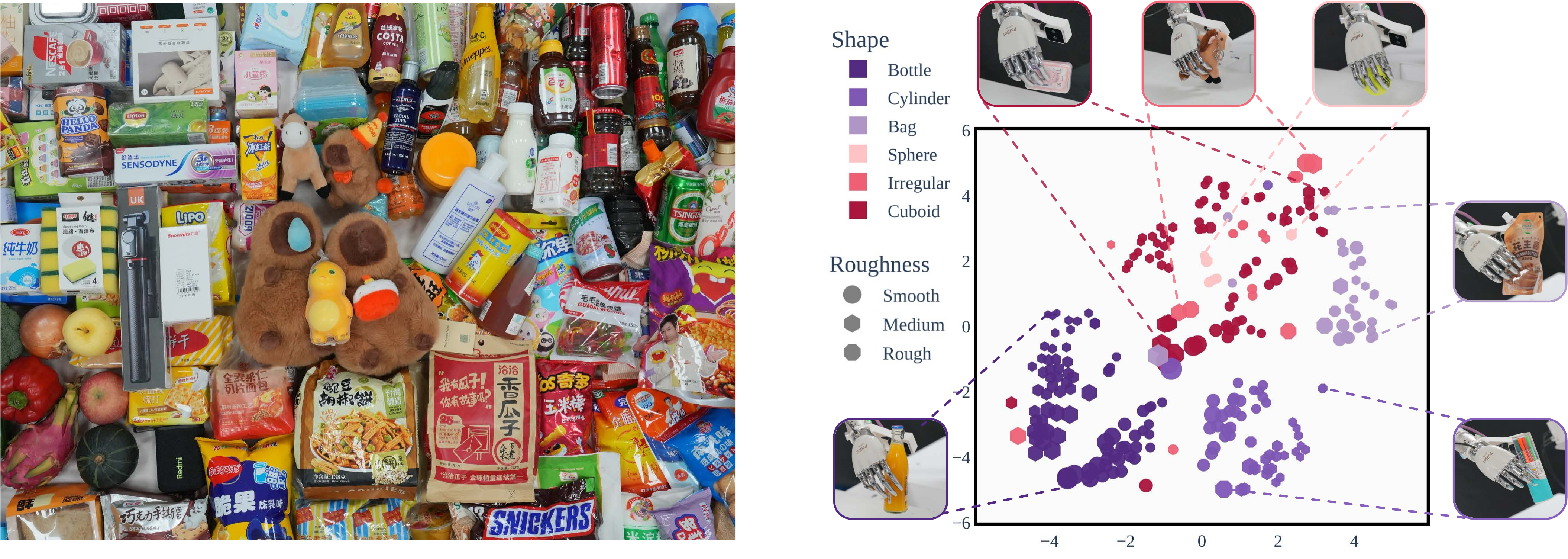}
  \caption{\textbf{(Left)} A representative part of all 360 unseen objects used to evaluate DexGraspVLA. \textbf{(Right)} A t-SNE projection illustrating the diversity and broad coverage of these objects in length, width, height, mass (denoted by marker size), roughness (marker type), and shape (marker color).}
  \label{fig:tSNE}
  \vspace{-15pt}
\end{figure*}

\begin{figure*}[t]
    \centering

    \begin{subfigure}[t]{0.4\linewidth}
        \centering
        \includegraphics[width=\linewidth]{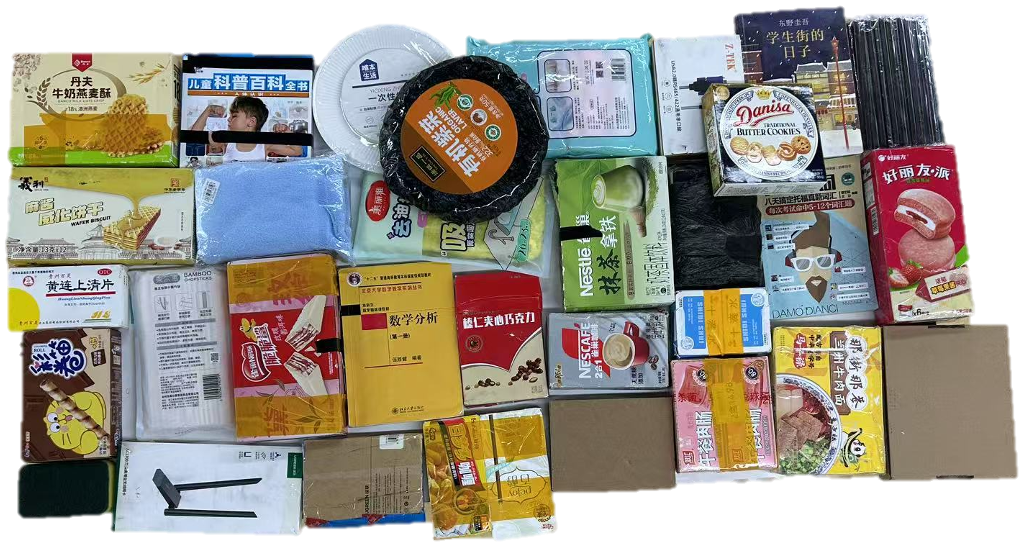}
        \captionsetup{justification=centering}
        \caption{The 32 objects used to collect nonprehensile grasping demonstrations.}
        \label{fig:nonprehensile-demo-objects}
    \end{subfigure}
    \hspace{2em}
    \begin{subfigure}[t]{0.4\linewidth}
        \centering
        \includegraphics[width=\linewidth]{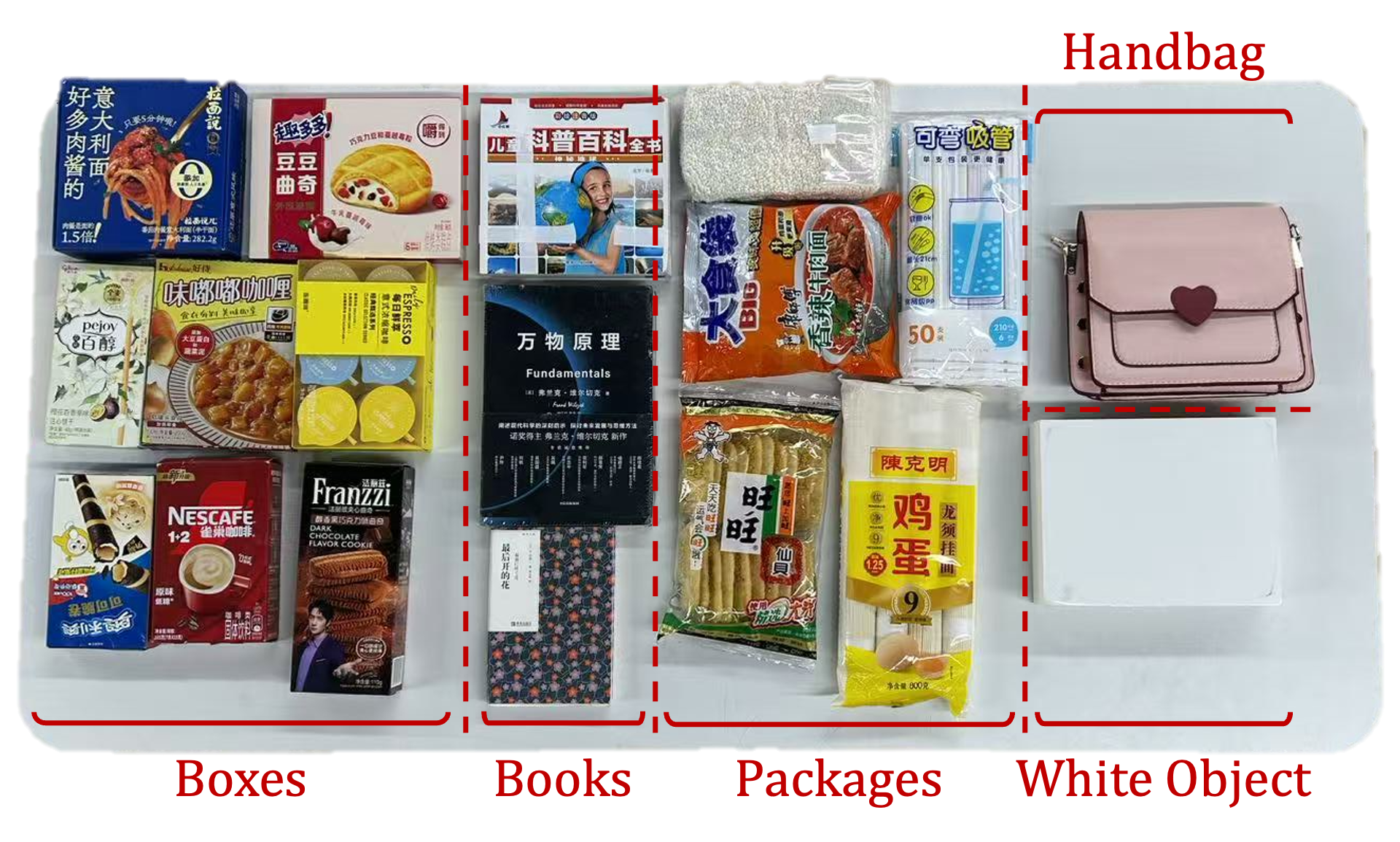}
        \captionsetup{justification=centering}
        \caption{The 18 objects used to test methods' performance in nonprehensile grasping.}
        \label{fig:nonprehensile-test-objects}
    \end{subfigure}

    \caption{Objects used to train and test methods in nonprehensile grasping. DexGraspVLA achieves robust generalization performance on diverse unseen objects.}
    \label{fig:nonprehensile-objects}
    \vspace{-10pt}
\end{figure*}

We collect demonstrations through kinesthetic teaching. At the beginning, the robot is set to teaching mode, allowing manual guidance to grasp target objects. The operator then physically guides the robot to the target position and performs the grasping motion.
Subsequently, we reset the environment and execute PD control using the recorded joint angles as target. At the same frequency, these target joint angles serve as actions, while images and current joint angles are collected as states. Following the same approach as the low-level controller, we post-process the collected data to generate masks, completing one demonstration sequence. In the general dexterous grasping experiments, each episode has a fixed duration of 75 timesteps, while in nonprehensile grasping, demonstrations have variable lengths, depending on the amount of manipulation required to push the object toward the table edge and complete the grasp. The control frequency is 20Hz.

We hire external contractors, provide them with training, and engage them to assist with data collection. All contractors were compensated with fair wages.

\section{Experiment Details}
\label{app:experiment-details}

\subsection{The ``Zero-Shot'' Evaluation Environment}

\Cref{fig:zero-shot-env} contrasts our data collection site and the test site, which are located in separate rooms. We gather all human demonstrations at the data collection site (\Cref{fig:data-collection-env}), whereas the experiments in \Cref{sec:experiments} are conducted at the test site (\Cref{fig:test-env}). Because these sites differ in layout and background, both the head camera and the wrist camera encounter scenes not present in the training data during evaluation --- particularly the wrist camera, which observes a notably altered environment, capturing a variety of front and peripheral views during operation. Despite these environmental discrepancies, we do not collect any data from the test site to fine-tune the models. Instead, the models are deployed and evaluated directly, resulting in a genuinely ``zero-shot'' testing environment. 
Even under these conditions, DexGraspVLA achieves an over 90\% success rate in grasping tasks in cluttered scenes across thousands of unseen object, lighting, and background combinations, clearly demonstrating its strong generalization capability.

\subsection{Additional Details of Objects, Lightings, and Backgrounds in General Dexterous Grasping}
\label{app:details-dexterous-grasping}

We collect 360 unseen objects with diverse sizes, weights, geometries, textures, materials, and categories. \Cref{fig:tSNE} presents the collected objects along with a t-SNE visualization of their measured properties, clearly demonstrating the high diversity of the object set. From these, 103 items are randomly selected as the \emph{object subset}~$\mathcal{S}$. In the large-scale generalization evaluation in the main paper (\Cref{subsec:generalization-eval}), the \emph{Unseen Objects} experiment is conducted on all 360 objects, while the \emph{Unseen Lightings} and \emph{Unseen Backgrounds} experiments use only the objects in $\mathcal{S}$. The three unseen lighting conditions comprise disco light, lamp light, and dark light. Meanwhile, the six unseen backgrounds include a black mouse pad, a pink towel, a colorful tablecloth, a black-and-white mouse pad, a wooden board, and a calibration board. These conditions are illustrated in \Cref{fig:generalization-eval-env}. In the baseline comparison experiments (\Cref{subsec:baseline-comparison}), the two unseen lighting conditions are disco light and lamp light, while the unseen backgrounds are a colorful tablecloth and a black-white mouse pad.

\subsection{Additional Details of Objects, Lightings, and Backgrounds in Nonprehensile Grasping}

In \Cref{fig:nonprehensile-objects}, we present the 32 objects curated for collecting nonprehensile grasping demonstrations and 18 unseen objects used for evaluation, covering a wide range of appearances, geometries, sizes, and categories. In \Cref{fig:generalization-eval-env}, we show the unseen background and lighting conditions used in the generalization evaluation. DexGraspVLA demonstrates robust performance on challenging cases, including fully white or irregularly shaped objects. In these scenarios, it successfully pushes the objects toward the table edge to enable stable grasping, even under complex and unseen lighting and background conditions.

\subsection{Details of Visualization}
\label{app:visualization-details}

\begin{figure}[t]
  \centering
  \includegraphics[width=\linewidth]{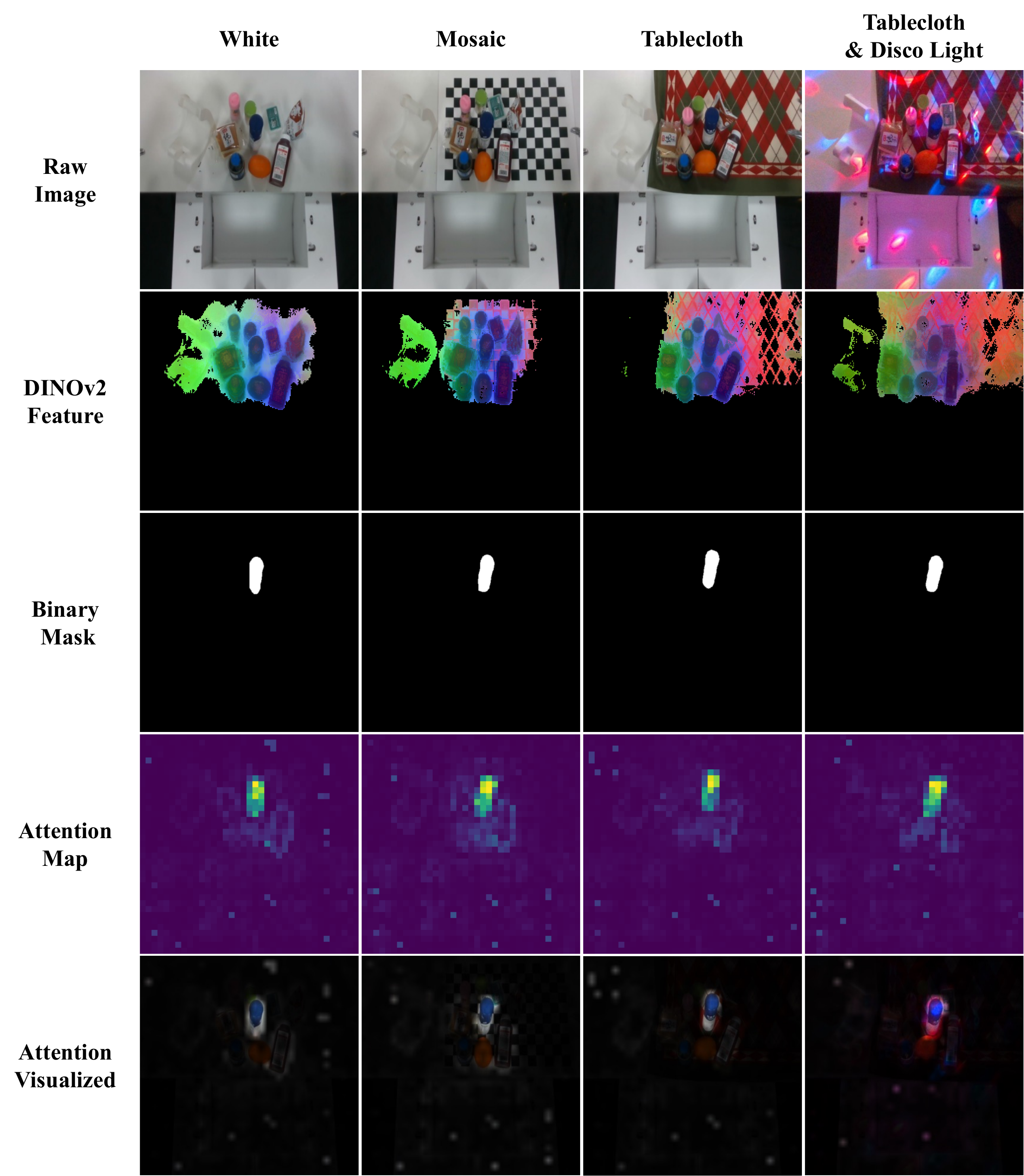}
  \captionsetup{justification=centering}
  \caption{The complete, uncropped version of \Cref{fig:dino-attn}.}
  \label{fig:dino-attn-full}
\end{figure}

In this part, we explain how we visualize the internal model behavior shown in \Cref{fig:dino-attn}. Due to space constraints, \Cref{fig:dino-attn} only presents the relevant portion of images containing the tabletop workspace. The full version is shown in \Cref{fig:dino-attn-full}. The first row is raw images from the head camera resized to $518\times518\times3$. The second row illustrates the DINOv2 ViT-B/14 features following the practice introduced in DINOv2 paper~\cite{oquab2023dinov2}. To make the resulting feature map recognizable for visualization purpose, we enlarge both the height and weight of images by a factor of six before feeding them into DINOv2. After obtaining the feature sequences for all four images, we combine these features, perform a PCA between all patches,  and set a threshold to remove background regions. We then apply PCA again, this time to the remaining foreground features, map the top three principal components to the RGB channels, and normalize the result. This yields the visualization shown in the second row. The third row showcases the binary masks $\mathbf{m}_t\in\mathbb{R}^{518\times518\times1}$ tracked by Cutie. The fourth row displays the averaged DiT attention maps over the head image features. This is computed by summing attention weights to each head image patch across all diffusion steps, DiT layers, DiT heads, and action tokens, and normalize the sum to one. The shape of the averaged attention map is $37 \times37\times1$. Finally, we upsample the attention map to $518 \times 518\times 1$, multiply it by 2 to increase brightness, and use it to scale the value channel of head images in HSV space, resulting in the visualization shown in the fifth row.

\section{Additional Results}
\label{app:additional-results}

This section provides additional results for the experiments in the main paper. In \Cref{tab:detail_success}, we report the detailed success rates for our large-scale generalization evaluation under each environment condition, corresponding to \Cref{tab:large-scale-generalization} in \Cref{subsec:generalization-eval}. From the first row (``Ours@1''), it is evident that DexGraspVLA maintains consistently high success rates across various unseen object, lighting, and background combinations. Many observed failures stem from randomness in policy inference; allowing additional attempts often recovers these failed cases. Accordingly, the second and third rows (“Ours@2” and “Ours@3”) show further improvements in performance, highlighting the potential for DexGraspVLA to reach even higher success rates. 

In \Cref{tab:detail_baselines}, we present the detailed success rates of DexGraspVLA and baseline methods on the smaller-scale dexterous grasping experiments described in \Cref{subsec:baseline-comparison}. Despite being evaluated in a zero-shot setting, DexGraspVLA consistently achieves success rates around or above 90\%. In contrast, baseline methods often fail catastrophically even on seen objects and perform even worse in unseen cluttered scenes. These results underscore the superior generalization capabilities of DexGraspVLA.

\begin{table*}[t]
  \caption{The detailed performance of DexGraspVLA under different unseen conditions, which indicates that our approach consistently achieves high success rates across various objects, lightings, and backgrounds. The second and third rows highlight its potential to reach even higher success rates given more chances. Numbers in parentheses indicate the number of tests.}
  \label{tab:detail_success}
  \centering
  \footnotesize
  \resizebox{\textwidth}{!}{
  \begin{tabular}{@{}c|c|ccc|cccccc|c@{}}
    \toprule
    Tasks      
    & \makecell{Unseen \\ Objects \\ ($360$)}
    & \multicolumn{3}{c|}{\makecell{Unseen Lightings \\ ($3\times103$)}}  
    & \multicolumn{6}{c|}{\makecell{Unseen Backgrounds \\ ($6\times103$)}}
    & \multirow{3}{*}{\makecell{Aggregated \\ ($1287$)}} \\
    
    \cmidrule(lr){1-11}
    
    \makecell{Lighting \\ Conditions} 
    & \makecell{White \\ Light}  
    & \makecell{Disco \\ Light}  
    & \makecell{Lamp \\ Light}  
    & \makecell{Dark \\ Light}  
    & \makecell{White \\ Light}  
    & \makecell{White \\ Light}  
    & \makecell{White \\ Light}  
    & \makecell{White \\ Light}  
    & \makecell{White \\ Light}  
    & \makecell{White \\ Light} 
    & \\
    
    \cmidrule(lr){1-11}

    \makecell{Background \\ Conditions}
    & \makecell{White \\ Table}  
    & \makecell{White \\ Table}  
    & \makecell{White \\ Table}  
    & \makecell{White \\ Table}  
    & \makecell{Black \\ Mouse Pad}  
    & \makecell{Pink \\ Towel}  
    & \makecell{Colorful \\ Tablecloth}  
    & \makecell{Black-White\\ Mouse Pad}  
    & \makecell{Wooden \\ Board}  
    & \makecell{Calibration \\ Board} 
    & \\
    
    \midrule

    Ours@1 & $91.1\%$ & $92.2\%$ & $89.3\%$ & $91.2\%$ & $94.2\%$ & $84.5\%$ & $90.3\%$ & $92.2\%$ & $93.2\%$ & $88.3\%$ & $90.8\%$  \\
    Ours@2 & $95.3\%$ & $97.0\%$ & $95.1\%$ & $93.2\%$ & $97.1\%$ & $90.3\%$ & $91.3\%$ & $95.1\%$ & $98.1\%$ & $93.2\%$ & $94.7\%$  \\
    Ours@3 & $96.7\%$ & $98.1\%$ & $98.1\%$ & $96.1\%$ & $98.1\%$ & $91.3\%$ & $94.2\%$ & $98.1\%$ & $100.0\%$ & $98.1\%$ & $96.9\%$ \\
    
    \bottomrule
  \end{tabular}
  }
\end{table*}

\begin{table*}[t]
 \caption{Performance of various methods on dexterous grasping in cluttered scenes under different conditions. DexGraspVLA consistently outperforms baselines, demonstrating superior generalization. Numbers in parentheses indicate the number of tests.}
 \label{tab:detail_baselines}
 \centering
 \footnotesize
 \begin{tabular}{@{}c|c|c|cc|cc|c@{}}
  \toprule
  Tasks
  & \makecell{Seen Objects \\ ($12$)}
  & \makecell{Unseen Objects \\ ($24$)}
  & \multicolumn{2}{c|}{\makecell{Unseen Lightings \\ ($2\times24$)}}
  & \multicolumn{2}{c|}{\makecell{Unseen Backgrounds \\ ($2\times24$)}}
  & \multirow{3}{*}{\makecell{Aggregated \\ ($132$)}} \\
  
  \cmidrule(lr){1-7}
  
  \makecell{Lighting \\ Conditions}
  & \makecell{White \\ Light}
  & \makecell{White \\ Light}
  & \makecell{Disco \\ Light}
  & \makecell{Lamp \\ Light}
  & \makecell{White \\ Light}
  & \makecell{White \\ Light}
  & \\
  
  \cmidrule(lr){1-7}
  
  \makecell{Background \\ Conditions}
  & \makecell{White \\ Table}
  & \makecell{White \\ Table}
  & \makecell{White \\ Table}
  & \makecell{White \\ Table}
  & \makecell{Black-White\\ Mouse Pad}
  & \makecell{Colorful \\ Tablecloth}
  & \\
  
  \midrule

  OpenVLA (LoRA)        & 33.0\% & 16.7\% & 16.7\% & 4.2\% & 12.5\% & 4.2\% & 12.9\% \\
  OpenVLA-OFT (LoRA)    & 33.0\% & 29.2\% & 20.8\% & 16.7\% & 41.7\% & 45.8\% & 30.3\% \\
  RDT (Full FT)         & 25.0\% & 25.0\% & 37.5\% & 41.6\% & 25.0\% & 29.2\% & 31.1\% \\
  $\pi_0$ (LoRA)        & 58.3\% & 45.8\% & 20.8\% & 16.7\% & 8.3\% & 4.1\% & 22.7\% \\
  $\pi_0$ (Full FT)     & 75.0\% & 45.8\% & 41.7\% & 25.0\% & 0.0\% & 16.7\% & 30.3\% \\
  Ours                  & \textbf{91.7\%} & \textbf{91.7\%} & \textbf{87.5\%} & \textbf{91.7\%} & \textbf{91.7\%} & \textbf{95.8\%} & \textbf{91.7\%} \\
  \bottomrule
 \end{tabular}
\end{table*}

\begin{figure*}[t]
    \centering
    \includegraphics[width=1.0\linewidth]{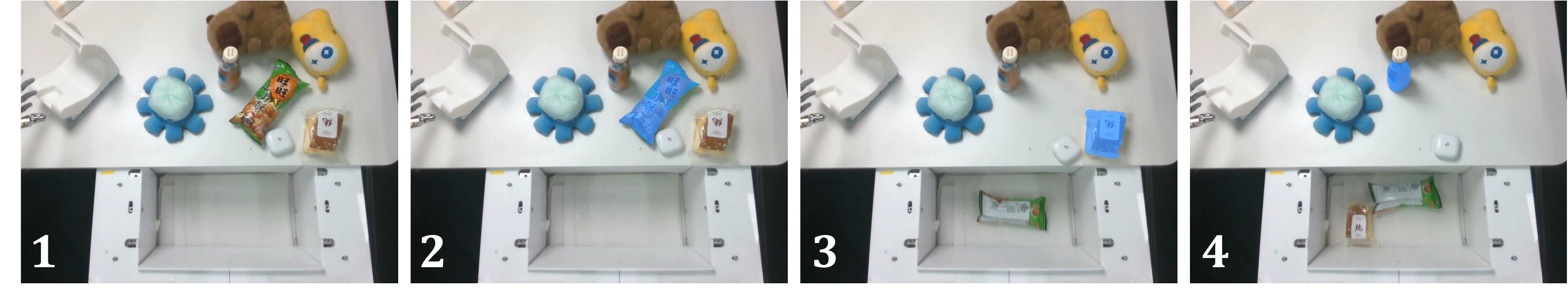}
    \caption{The head camera images observed by DexGraspVLA planner and generated masks for a long-horizon task: "Grasp all edible objects, including food and drinks."}
    \label{fig:long-horizon-initial}
\end{figure*}

We next present the generated responses of DexGraspVLA planner in an exemplar long-horizon task, which demonstrates visual understanding and embodied reasoning. The cluttered scene is arranged as shown in \Cref{fig:long-horizon-initial}, with the user prompt being "Grasp all edible objects, including food and drinks". Below is the generated responses of DexGraspVLA planner. We omit our system prompts as they have been elaborated in \Cref{app:dexgraspvla-details}.

\begin{tcolorbox}[breakable]

\small

<System Prompt of Prompt Completion Check>

......
\newline\newline
<Planner Response>

To determine if the user prompt has been fully completed, let's identify all the edible objects in the initial image and check if they have been removed or grasped in the current image.
\newline\newline
\#\#\# Edible Objects in the Initial Image:

1. A bottle (drink).

2. A green packet with an orange label (food).

3. A small white object with a heart symbol (possibly food).

4. A packaged item with a brown appearance (food).
\newline\newline
\#\#\# Analysis of Current Image:

- The bottle is still present on the table.

- The green packet with an orange label is still present on the table.

- The small white object with a heart symbol is still present on the table.

- The packaged item with a brown appearance is still present on the table.
\newline\newline
Since all the edible objects are still visible on the table in the current image, none of them have been removed or grasped.
\newline\newline
**Reasoning:** All target objects (edible items) remain on the table, indicating that the user prompt to grasp all edible objects has not been completed.
\newline\newline
**Output:** False
\newline\newline
<System Prompt of Instruction Proposal>

......
\newline\newline
<Planner Response>

The green and orange snack bag on the right side of the table.
\newline\newline
<System Prompt of Bounding Box Prediction>

......
\newline\newline
<Planner Response>

\begin{lstlisting}[basicstyle=\ttfamily, breaklines=true, numbers=none]
{
    "bbox_2d": [357, 104, 486, 249],
    "label": "green orange snack",
    "description": "The green and orange snack bag on the right side of the table."
}
\end{lstlisting}

<System Prompt of Grasp Outcome Verification>

......
\newline\newline
<Planner Response>

The target object is no longer visible on the table and appears to be securely held in the robotic hand.
\newline\newline
True
\newline\newline
<System Prompt of Prompt Completion Check>

......
\newline\newline
<Planner Response>

To determine if the user prompt has been fully completed, let's identify all the edible objects in the initial image and check if they have been removed or grasped in the current image.
\newline\newline
\#\#\# Initial Image Edible Objects:

1. A bottle (drink).

2. A green packet with a brown label (food).

3. A small packaged food item with a white label (food).
\newline\newline
\#\#\# Current Image Analysis:

- The bottle is still present on the table.

- The green packet with a brown label is no longer visible.

- The small packaged food item with a white label is still present on the table.
\newline\newline
\#\#\# Reasoning:
The user prompt was to grasp all edible objects. However, the bottle and the small packaged food item are still on the table, indicating that not all edible objects have been removed or grasped.
\newline\newline
\#\#\# Conclusion:
Since some relevant objects remain on the table, the user prompt has not been fully completed.
\newline\newline
**Output:**
All edible objects have been removed or grasped: False.
\newline\newline
<System Prompt of Instruction Proposal>

......
\newline\newline
<Planner Response>

The brown snack packet on the right side of the table.
\newline\newline
<System Prompt of Bounding Box Prediction>

......
\newline\newline
<Planner Response>

\begin{lstlisting}[basicstyle=\ttfamily, breaklines=true, numbers=none]
{
    "bbox_2d": [450, 168, 579, 263],
    "label": "brown snack packet",
    "description": "The brown snack packet on the right side of the table"
}
\end{lstlisting}

<System Prompt of Grasp Outcome Verification>

......
\newline\newline
<Planner Response>

The target object is not visible on the table, and it appears to be securely held in the robotic hand based on the close-up view.
\newline\newline
Reasoning: The object is no longer on the table and is being held by the robotic arm.

True
\newline\newline
<System Prompt of Prompt Completion Check>

......
\newline\newline
<Planner Response>

The user prompt is to grasp all edible objects, including food and drinks. In the initial image, there are several items that could be considered edible:
\newline\newline
1. A bottle of liquid (drink).

2. A bag of snacks.

3. A packaged food item.
\newline\newline
In the current image, the bottle of liquid remains on the table, while the other two items (the bag of snacks and the packaged food) are no longer visible. This suggests that only some of the edible objects have been grasped or removed.
\newline\newline
Reasoning: Since the bottle of liquid is still present on the table, not all edible objects have been grasped.
\newline\newline
Output: False
\newline\newline
<System Prompt of Instruction Proposal>

......
\newline\newline
<Planner Response>

The white bottle with a brown label next to the yellow toy on the right side of the table.
\newline\newline
<System Prompt of Bounding Box Prediction>

......
\newline\newline
<Planner Response>

\begin{lstlisting}[basicstyle=\ttfamily, breaklines=true, numbers=none]
{
    "bbox_2d": [304, 150, 368, 227],
    "label": "white bottle",
    "description": "The white bottle with a brown label is located next to the yellow toy on the right side of the table."
}
\end{lstlisting}

<System Prompt of Grasp Outcome Verification>

......
\newline\newline
<Planner Response>

The target object is no longer visible on the table and appears to be securely held in the robotic hand.
\newline\newline
True
\newline\newline
<System Prompt of Prompt Completion Check>

......
\newline\newline
<Planner Response>

The user prompt is to grasp all edible objects, including food and drinks. In the initial image, there are several items that could be considered edible:
\newline\newline
1. A bottle of liquid (possibly a drink).

2. A packaged snack with visible text.

3. Another packaged item that appears to be food.
\newline\newline
In the current image, these items are no longer visible on the table. This suggests they have been removed or grasped by the robot.
\newline\newline
Reasoning: Since all the potential edible objects from the initial image are no longer present in the current image, it can be concluded that the user prompt has been fully completed.
\newline\newline
Output: True

\end{tcolorbox}

\section{Additional Experiments}
\label{app:additional-experiments}

This section presents an additional experiment that aims to separately evaluate the bounding-box prediction accuracy of the DexGraspVLA planner.

\begin{figure}[t]
    \centering
    \begin{subfigure}[b]{0.6\linewidth}
        \centering
        \includegraphics[width=\linewidth]{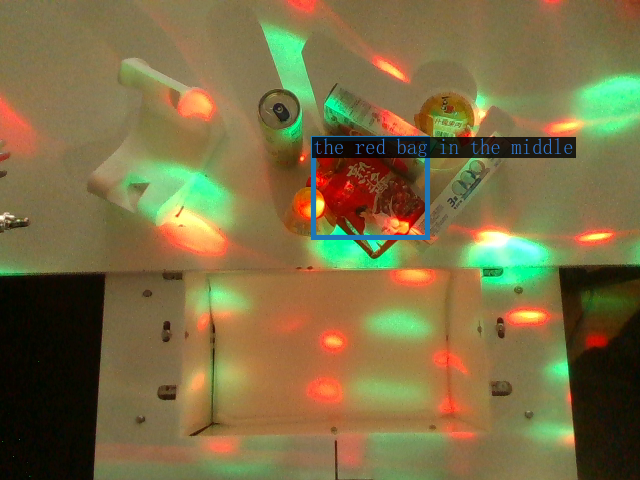}
    \end{subfigure}
    \vspace{1pt}
    \begin{subfigure}[b]{0.6\linewidth}
        \centering
        \includegraphics[width=\linewidth]{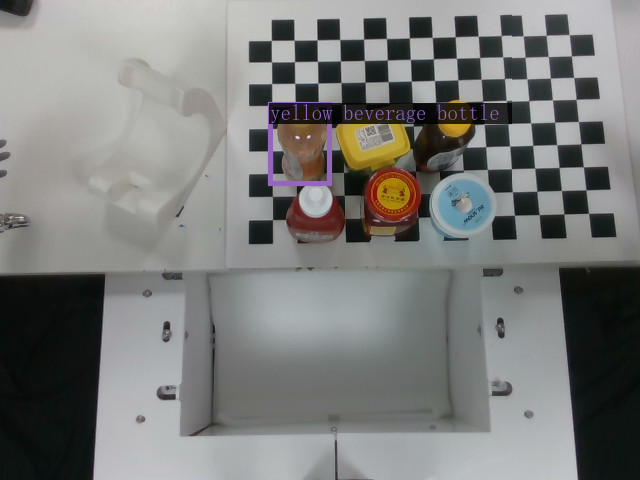}
    \end{subfigure}
    \vspace{1pt}
    \begin{subfigure}[b]{0.6\linewidth}
        \centering
        \includegraphics[width=\linewidth]{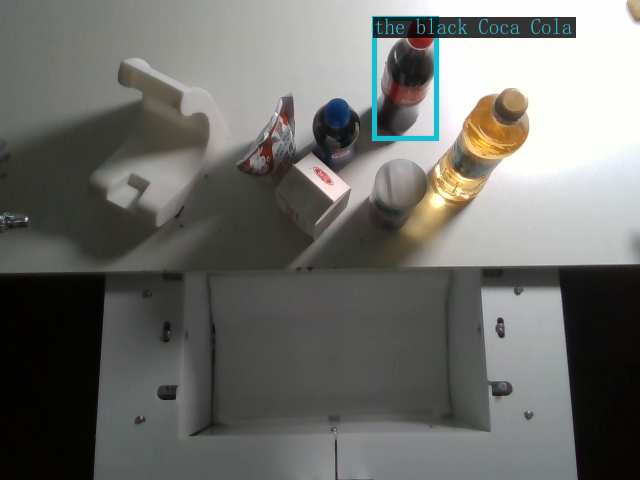}
    \end{subfigure}
    \vspace{1pt}
    \begin{subfigure}[b]{0.6\linewidth}
        \centering
        \includegraphics[width=\linewidth]{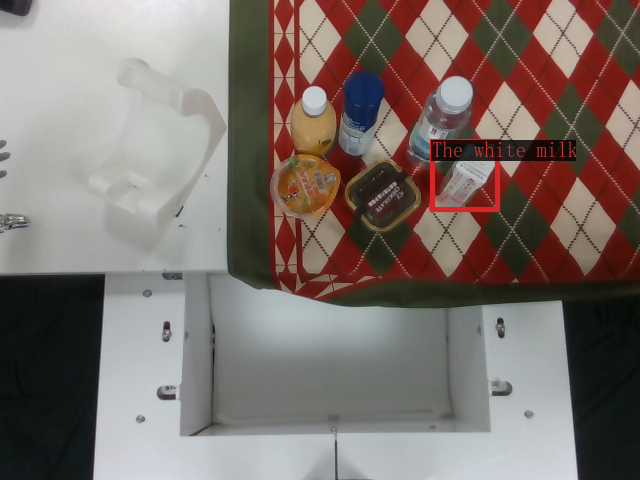}
    \end{subfigure}
    
    \caption{Bounding-box predictions made by DexGraspVLA planner. Across diverse lighting and background conditions, it accurately grounds the language instruction to the target object in cluttered scenes and marks the correct bounding box.}
    \label{fig:bbox}
\end{figure}

\paragraph{Tasks.} The bounding-box prediction accuracy of the planner is crucial to the success of grasping, as it determines the target for the controller. To evaluate this accuracy, we design three types of tasks featuring different environmental distractions: (1) \emph{No Distraction} (1 scenario): The cluttered scene is arranged on a white table under white light; (2) \emph{Background Distraction} (2 scenarios): The cluttered scene is placed on either a calibration board or a brightly colored tablecloth, both under white light; (3) \emph{Lighting Distraction} (2 scenarios): The scene is set up in a dark room illuminated by either a desk lamp or a disco light. Scenarios with distractions are shown in \Cref{fig:bbox}. For each scenario, we randomly arrange five cluttered scenes, each containing six randomly selected objects, and then record head-camera images. For each object, we provide a textual prompt describing its appearance and location, and check whether the planner’s bounding-box prediction accurately marks the target. In total, \emph{No Distraction} accounts for 30 tests, while \emph{Background Distraction}  and  \emph{Lighting Distraction} both have 60 tests, amounting to 150 tests overall.

\paragraph{Metric.} We define a bounding box as accurate if it tightly encloses the target object. Accuracy is then measured as the proportion of accurate bounding boxes over all tested objects.

\paragraph{Results.} The accuracy is reported in \Cref{tab:planner-accuracy}. For 150 prompts, the planner only mislabels one bounding box while succeeding in the other 149 tests, resulting in an aggregated accuracy exceeding 99\%. In \Cref{fig:bbox}, we present examples of bounding-box predictions produced by the DexGraspVLA planner. Despite substantial variation in environmental conditions, the planner consistently grounds grasping instructions in cluttered scenes and provides the correct bounding boxes. Notably, we can identify objects by names such as “Coca Cola” or “milk,” reflecting the system’s extensive common sense and world knowledge. By drawing on the broad knowledge embedded in each of its foundation models, DexGraspVLA achieves robust generalization across diverse scenarios.

\begin{table}[t]
\centering
\scriptsize
\setlength{\tabcolsep}{4pt}
\renewcommand{\arraystretch}{1.1}
\caption{Planner accuracy in bounding-box prediction under different environment conditions.}
\label{tab:planner-accuracy}
\begin{tabular}{lcccc}
  \toprule
  & \begin{tabular}{@{}c@{}}No\\Distraction\end{tabular}
  & \begin{tabular}{@{}c@{}}Background\\Distraction\end{tabular}
  & \begin{tabular}{@{}c@{}}Lighting\\Distraction\end{tabular}
  & \begin{tabular}{@{}c@{}}Aggregated\end{tabular} \\
  \midrule
  Planner & 96.7\% & 100.0\% & 100.0\% & 99.3\% \\
  \bottomrule
\end{tabular}
\end{table}

\end{document}